%% file: paper.tex
\documentclass[10pt,twocolumn,letterpaper]{article}

\usepackage[pagenumbers]{cvpr} % To force page numbers, e.g. for an arXiv version

\usepackage{graphicx}
\usepackage{amsmath}
\usepackage{amssymb}
\usepackage{booktabs}

\usepackage{threeparttable}
\usepackage{stmaryrd}
\usepackage{makecell}
\usepackage{marvosym}
\usepackage{xcolor}
\usepackage{enumerate}

\usepackage{algorithmicx,algorithm}
\usepackage{algpseudocode}
\usepackage{colortbl}
\definecolor{tbgray}{gray}{.9}

\usepackage{bm}

\usepackage[accsupp]{axessibility}

\usepackage[pagebackref,breaklinks,colorlinks]{hyperref}

\usepackage[capitalize]{cleveref}
\crefname{section}{Sec.}{Secs.}
\Crefname{section}{Section}{Sections}
\Crefname{table}{Table}{Tables}
\crefname{table}{Tab.}{Tabs.}
\def\aka{\emph{a.k.a}\onedot}

\def\cvprPaperID{4392} % *** Enter the CVPR Paper ID here
\def\confName{CVPR}
\def\confYear{2022}

\newcommand{\HL}[1]{\textcolor[rgb]{0.8,0.31,0.2235 }{\textbf{#1}}}
\newcommand{\gray}[1]{\textcolor[rgb]{0.753, 0.753, 0.753}{#1}}
\newcommand{\tabincell}[2]{\begin{tabular}{@{}#1@{}}#2\end{tabular}}

\begin{document}

\title{Safe Self-Refinement for Transformer-based Domain Adaptation}

\author{Tao Sun$^1$, Cheng Lu$^2$, Tianshuo Zhang$^2$, Haibin Ling$^1$\\
$^1$Stony Brook University, $^2$XPeng Motors\\
{\tt\small \{tao,hling\}@cs.stonybrook.edu, luc@xiaopeng.com, tonyzhang2035@gmail.com}
}
\maketitle

%%%%%%%%% ABSTRACT
\begin{abstract}
   Unsupervised Domain Adaptation (UDA) aims to leverage a label-rich source domain to solve tasks on a related unlabeled target domain. It is a challenging problem especially when a large domain gap lies between the source and target domains. In this paper we propose a novel solution named SSRT (Safe Self-Refinement for Transformer-based domain adaptation), which brings improvement from two aspects. First, encouraged by the success of vision transformers in various vision tasks, we arm SSRT with a transformer backbone. We find that the combination of vision transformer with simple adversarial adaptation surpasses best reported Convolutional Neural Network (CNN)-based results on the challenging DomainNet benchmark, showing its strong transferable feature representation. Second, to reduce the risk of model collapse and improve the effectiveness of knowledge transfer between domains with large gaps, we propose a Safe Self-Refinement strategy. Specifically, SSRT utilizes predictions of perturbed target domain data to refine the model. Since the model capacity of vision transformer is large and predictions in such challenging tasks can be noisy, a safe training mechanism is designed to adaptively adjust learning configuration. Extensive evaluations are conducted on several widely tested UDA benchmarks and SSRT achieves consistently the best performances, including 85.43\% on Office-Home, 88.76\% on VisDA-2017 and 45.2\% on DomainNet.
\end{abstract}

%%%%%%%%% BODY TEXT
\section{Introduction}
\label{sec:intro}

Deep neural networks have achieved impressive performance in a variety of machine learning tasks. However, the success often relies on a large amount of labeled training data, which can be costly or impractical to obtain. Unsupervised Domain Adaptation (UDA)~\cite{wilson2020survey} handles this issue by transferring knowledge from a label-rich source domain to a different unlabeled target domain.  Over the past years, many UDA methods have been proposed~\cite{ganin2015unsupervised,long2018conditional,zhang2019bridging,shu2018dirt,liang2020we}. Among them, adversarial adaptation~\cite{ganin2015unsupervised,long2018conditional,zhang2019bridging} that learns domain-invariant feature representation using the idea of adversarial learning has been a prevailing paradigm. Deep UDA methods are usually applied in conjunction with a pretrained Convolutional Neural Network (CNN, \eg, ResNet~\cite{he2016deep}) backbone in vision tasks. On medium-sized classification benchmarks such as Office-Home~\cite{venkateswara2017deep} and VisDA~\cite{peng2017visda}, the reported state-of-the-arts are very impressive~\cite{liang2020we}. However, on large-scale datasets like DomainNet~\cite{peng2019moment}, the most recent results in the literature by our submission report a best average accuracy of 33.3\%~\cite{li2021semantic}, which is far from satisfactory. 

With the above observations, we focus our investigation on challenging cases from two aspects: 
\begin{enumerate}[$\bullet$]
\vspace{-1.6mm}\item {First}, from the \textit{representation aspect}, it is desirable to use a more powerful backbone network. This directs our attention to the recently popularized vision transformers, which have been successfully applied to various vision tasks~\cite{dosovitskiy2020image,carion2020end,zhang2021multi}. Vision transformer processes an image as a sequence of tokens, and uses global self-attention to refine this representation. With its long-range dependencies and large-scale pre-training, vision transformer obtains strong feature representation that is ready for down-stream tasks. Despite this, its application in UDA is still under-explored. Hence we propose to integrate vision transformer to UDA. We find that by simply combining  ViT-B/16~\cite{dosovitskiy2020image} with adversarial adaptation, it can achieve 38.5\% average accuracy on DomainNet, better than the current arts using  ResNet-101\cite{he2016deep,li2021semantic}. This shows that the  feature representation of vision transformer is discriminative as well as transferable across domains.

\vspace{-1.6mm}\item {Second}, from the \textit{domain adaptation aspect}, a more reliable strategy is needed to protect the learning process from collapse due to large domain gaps. As strong backbones with large capacity like vision transformer increase the chance of overfitting to source domain data, a regularization from target domain data is desired. A common practice in UDA is to utilize model predictions for self-training or enforce clustering structure on target domain data~\cite{zhang2020label,shu2018dirt,liang2020we}. While this helps generally, the supervisions can be noisy when the domain gap is large. Therefore, an adaptation method is expected to be \emph{Safe}~\cite{li2014towards} enough to avoid model collapse.
\end{enumerate}

Motivated by the above discussions, in this paper, we propose a novel UDA solution named SSRT (Safe Self-Refinement for Transformer-based domain adaptation). SSRT takes a vision transformer as the backbone network and utilizes predictions on perturbed target domain data to refine the adapted model. Specifically, we add random offsets to the latent token sequences of target domain data, and minimize the discrepancy of model's predicted probabilities between the original and perturbed versions using the Kullback Leibler (KL) divergence. This imposes a regularization on the corresponding transformer layers in effect. Moreover, SSRT has several important components that contribute to its excellent performance, including multi-layer perturbation and bi-directional supervision.

To protect the learning process from collapse, we propose a novel \emph{Safe Training} mechanism. As UDA tasks vary widely even when they are drawn from the same dataset, a specific learning configuration (\eg, hyper-parameters) that works on most tasks may fail on some particular ones. The learning configuration is thus desired to be automatically adjustable. For example, for perturbation-based methods~\cite{miyato2018virtual,sohn2020fixmatch}, a small perturbation may under-exploit their benefits while a large one may result in collapse. Recent works~\cite{tarvainen2017mean, berthelot2019mixmatch} apply a manually defined ramp-up period at the beginning of training. However, this cannot solve the issue when its maximum value is improper for the current task. In contrast, we propose to monitor the whole training process and adjust learning configuration adaptively. We use a diversity measure of model predictions on the target domain data to detect model collapse. Once it occurs, the model is restored to a previously achieved state and the configuration is reset. With this safe training strategy, our SSRT avoids significant performance deterioration on adaptation tasks with large domain gaps. The code is available at \url{https://github.com/tsun/SSRT}.

In summary, we make the following contributions:
\begin{enumerate}[$\bullet$]
	\vspace{-1.6mm}\item We develop a novel UDA solution SSRT, which adopts a vision transformer backbone for its strong transferable feature representation, and utilizes the predictions on perturbed target domain data for model refinement.
	\vspace{-1.6mm}\item We propose a safe training strategy to protect the learning process from collapse due to large domain gaps. It adaptively adjusts learning configuration during the training process with a diversity measure of model predictions on target domain data.
	\vspace{-1.6mm}\item SSRT is among the first to explore vision transformer for domain adaptation. Vision transformer-based UDA has shown promising results, especially on large-scale datasets like DomainNet. 
	\vspace{-1.6mm}\item Extensive experiments are conducted on widely tested benchmarks. Our SSRT achieves the best performances, including 85.43\% on Office-Home, 88.76\% on VisDA-2017 and 45.2\% on DomainNet.
\end{enumerate}

\begin{figure*}[!t]
	\begin{center}
			\centering
			\includegraphics[width=0.97\linewidth]{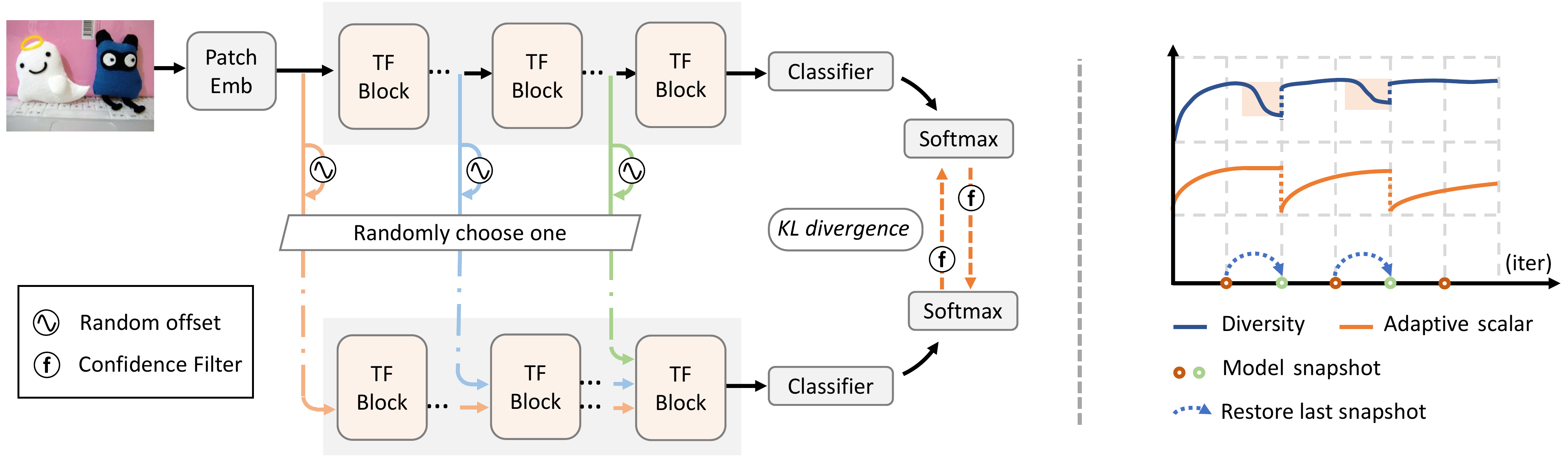}
	\end{center}
	\caption{Overview of SSRT. (\textbf{Left}) Illustration of Self-Refinement for our transformer-based model. The two branches share parameters. Random offsets are added to the input token sequences of transformer (TF) blocks. The model is refined using its predictions of the original and perturbed versions supervised by KL divergence. (\textbf{Right}) Illustration of Safe Training mechanism. See text for details.		
 }
	\label{fig:SD}
\end{figure*}

\section{Related Work}
\vspace{-1mm}
\paragraph{Unsupervised Domain Adaptation.} There are several prevailing categories of UDA methods. Discrepancy-based methods minimize the distribution divergence between source and target domains with discrepancy measures~\cite{tzeng2014deep,long2017deep,sun2016deep}. Adversarial adaptation methods learn domain-invariant representations by playing a two-player min-max game between the feature extractor and a domain discriminator~\cite{tzeng2014deep,sun2016deep,ganin2015unsupervised,tzeng2017adversarial}. Recently, many works exploit self-training for domain adaptation~\cite{mei2020instance,zou2019confidence,zou2018unsupervised}. They generate pseudo labels for target domain data and take them as labeled data to refine the model. 

\vspace{-2.5mm}\paragraph{Transformer in Vision.}
Vision Transformer (ViT)~\cite{dosovitskiy2020image} is a pioneering work that applies a convolution-free transformer structure for image classification. Following that, many ViT variants have been proposed~\cite{touvron2021training,yin2020disentangled,han2021transformer,liu2021swin}. Transformer has been applied successfully to various vision tasks including image classification~\cite{dosovitskiy2020image,touvron2021training}, object detection~\cite{carion2020end}, semantic segmentation~\cite{strudel2021segmenter}, \etc.  

The application of vision transformer in domain adaptation, however, is still very scarce. Notably, two concurrent explorations~\cite{xu2021cdtrans,yang2021tvt} have been recently reported on arXiv. Specifically, CDTrans~\cite{xu2021cdtrans} is a pure transformer solution for UDA, and it applies cross attention on source-target image pairs. TVT~\cite{yang2021tvt} proposes a transferable multi-head self-attention module and combines it with adversarial adaptation. Our method is different in that it uses pairs of target domain data and their perturbed version to refine the model. This guarantees the same semantic class. Besides, we delicately design the components of our model and the training strategy to avoid collapse on challenging tasks.

\vspace{-2.5mm}\paragraph{Consistency Regularization.} Consistency regularization is an important technique in semi-supervised learning that achieves state-of-the-art results~\cite{sohn2020fixmatch}. It leverages the idea that model predictions should be similar for semantically identical data. Some methods create perturbed inputs with adversarial training~\cite{miyato2018virtual}, while others use standard data augmentations~\cite{berthelot2019mixmatch,sohn2020fixmatch,xie2019unsupervised}. These works mostly manipulate raw input images. In contrast, our study focuses on the latent token sequence representation of vision transformer. 

%In addition, we use bi-directional supervision to refine the model, which proves to be more robust than uni-directional one as shown in our experiments. 

\section{Proposed Method}
\subsection{Problem Formulation}
\label{sec:background}
In Unsupervised Domain Adaptation, there is a source domain with labeled data $\mathcal{D}_s=\{(\bm{x}_i^s, y_i^s)\}_{i=1}^{n_s}$ from $\mathcal{X}\times \mathcal{Y}$ and a target domain with unlabeled data $\mathcal{D}_t=\{(\bm{x}_i^t)\}_{i=1}^{n_t}$ from $\mathcal{X}$, where $\mathcal{X}$ is the input space and $\mathcal{Y}$ is the label space. UDA aims to learn a classifier $h=g\circ f$, where $f(\cdot;\theta_f):\mathcal{X}\rightarrow \mathcal{Z}$ denotes the feature extractor, $g(\cdot;\theta_g):\mathcal{Z}\rightarrow \mathcal{Y}$ denotes the class predictor, and $\mathcal{Z}$ is the latent space. Adversarial adaptation learns domain-invariant feature via a 
 binary domain discrimination $d(\cdot;\theta_d):\mathcal{Z}\rightarrow [0,1]$ that maps features to domain labels. 
 
The objective is formulated as
\begin{equation}
 	\min_{f,g}\max_{d} \mathcal{L} = \mathcal{L}_{\mathrm{CE}} -\mathcal{L}_{\mathrm{d}} + \beta \mathcal{L}_{\mathrm{tgt}},
 	\label{eq:opt}
 \end{equation}
where $\mathcal{L}_{\mathrm{CE}}$ is the standard cross-entropy loss on source domain data, $\mathcal{L}_{\mathrm{d}}$ is domain adversarial loss, defined as 
 \begin{equation}
 	\begin{aligned}
 		\mathcal{L}_{\mathrm{d}} \!= \!-\mathbb{E}_{\bm{x}\sim \mathcal{D}_s}\!\big[\log d(f(\bm{x}))\big] \!-\! \mathbb{E}_{x\sim \mathcal{D}_t}\!\big[\log (1-d(f(\bm{x})))\big], 
 	\end{aligned}
 	\nonumber %\label{eq:dl}
 \end{equation}
$\beta$ is a trade-off parameter, and $\mathcal{L}_{\mathrm{tgt}}$ is a loss on target domain data. A common choice of $\mathcal{L}_{\mathrm{tgt}}$ is the Mutual Information Maximization loss~\cite{gomes2010discriminative,shi2012information}. In our method, we instantiate it as the self-refinement loss $\mathcal{L}_{\mathrm{SR}}$, introduced in Sec.~\ref{sec:sd}. 

\subsection{Method Framework}
We aim to regularize the latent feature spaces of transformer backbone by refining the model with perturbed target domain data. Figure~\ref{fig:SD} illustrates the framework of our proposed SSRT. Only target domain data are shown here. The network consists of a vision transformer backbone and a classifier head. Domain discriminator is not plotted. For each target domain image, the \emph{Patch Embedding} layer transforms it into a token sequence including a special class token and image tokens. Then the sequence is refined with a series of \emph{Transformer Blocks}. The classifier head takes the class token and outputs label prediction. We randomly choose one transformer block and add a random offset to its input token sequence. Then the corresponding predicted class probabilities of original and perturbed versions are used for bi-directional self-refinement. To avoid noisy supervision, only reliable predictions are used via a \emph{Confidence Filter}. To reduce the risk of model collapse, we use a safe training mechanism to learn the model.

\subsection{Multi-layer Perturbation for Transformer}

While many works manipulate the raw input images~\cite{miyato2018virtual,berthelot2019mixmatch,sohn2020fixmatch}, it may be better to do that at hidden layers~\cite{verma2019manifold}. Vision transformer has some particular properties due to its special architecture. Since the \emph{Patch Embedding} layer is merely a convolutional layer plus the position embedding, a linear operation on raw input can be shifted equivalently to the first transformer block. Besides, due to residual connections within transformer blocks, the token sequences at adjacent blocks are highly correlated. The best layer to add perturbation, however, varies across tasks. Empirically, perturbing relatively deep layers performs better but at a higher risk of model collapse. Therefore, we randomly choose one from multiple layers, which proves to be more robust than perturbing any single layer from them. In fact, it imposes a regularization on multiple layers simultaneously, making the learning process safer.

% \in \mathbb{R}^{t\times e}, where $t$ is the number of tokens and $e$ is the token embedding dimension. 
 
Given a target domain image $\bm{x}$, let $b^{l}_x$ be its input token sequence of the $l$-th transformer block. $b^{l}_x$ can be viewed as a latent representation of $\bm{x}$ in a hidden space. Since its dimension is high while the support of target domain data is limited in the space, it is inefficient to perturb $b^{l}_x$ arbitrarily. Instead, we utilize the token sequence $b^l_{xr}$ of another randomly chosen target domain image $\bm{x}_r$ to add an offset. The perturbed token sequence of $b^l_x$ is obtained as 
\begin{equation}
	\tilde{b}^{l}_x=b^{l}_x + \alpha [b_{xr}^{l}-b^{l}_x]_{\times},
	\label{eq:perturb}
\end{equation}   
where  $\alpha$ is a scalar and $[\cdot]_{\times}$ means no gradient back-propagation. Note that although gradients cannot back-propagate through the offset, they can pass through $b^{l}_x$. The importance of this is elaborated in the following section.

% Moreover, the gradient with respect to $\tilde{b}^{l}_x$ is propagated to $b^{l}_x$ identically.

In addition to the manually injected perturbation, the \emph{Dropout} layer in the classifier head also works randomly for the two branches. This creates another source of discrepancy for the self-refinement loss.

\subsection{Bi-directional Self-Refinement}
\label{sec:sd}

Now we are ready to define the loss function used for self-refinement. Let $\bm{p}_x$ and $\bm{\tilde{p}}_x$ be the predicted probability vectors corresponding to $b^{l}_x$ and $\tilde{b}^{l}_x$, respectively. To measure their distance, KL divergence is commonly used.
\begin{equation}
	D_{\textrm{KL}}(\bm{p}_t \| \bm{p}_s) = \sum_{i} \bm{p}_t[i]  \mathrm{log}\frac{\bm{p}_t[i]}{\bm{p}_s[i]} ,
	\label{eq:kl}
\end{equation}
where $\bm{p}_t$ is the teacher probability (\aka target probability) and $\bm{p}_s$ is the student probability. Note that KL divergence is asymmetric in $\bm{p}_t$ and $\bm{p}_s$. While it is natural to take $\bm{p}_x$ as the teacher probability since it corresponds to the original data, we find the reverse also works. Moreover, as shown in Sec.~\ref{sec:bisup}, it is more robust to combine them together. Our bi-directional self-refinement loss is defined as 
\begin{equation}
	\begin{aligned}
		\mathcal{L}_{\mathrm{SR}} =	\mathbb{E}_{\mathcal{B}_t\sim \mathcal{D}_t} &\Big\{\omega \mathbb{E}_{\bm{x}\sim F[\mathcal{B}_t;\bm{p}]}  D_{\textrm{KL}}(\bm{p}_x \| \bm{\tilde{p}}_x)\\
		& + (1-\omega) \mathbb{E}_{\bm{x}\sim F[\mathcal{B}_t;\bm{\tilde{p}}]}   D_{\textrm{KL}}(\bm{\tilde{p}}_x \| \bm{p}_x)\Big\} ,
	\end{aligned}
	\label{eq:SD}
\end{equation}
where $\omega$ is a random variable drawn from a Bernoulli distribution $\mathcal{B}(0.5)$, $F$ is a \emph{Confidence Filter} defined as 
\begin{equation}
	%F[p]=\llbracket \mathrm{max}(p)>\epsilon \rrbracket
	F[\mathcal{D};\bm{p}]=\{\bm{x}\in{\mathcal{D}}|\mathrm{max}(\bm{p}_x)>\epsilon\},
\end{equation}
and $\epsilon$ is a predefined threshold. $\mathcal{L}_{\mathrm{SR}}$ refines the model with confident predictions and regularizes it to predict smoothly in the latent feature spaces. 

Typically, the loss gradient is only back-propagated through the student probability (\ie, $\bm{p}_s$ in Eq.~\ref{eq:kl})~\cite{miyato2018virtual,oliver2018realistic,berthelot2019mixmatch}. We find, however, it is better to back-propagate gradient through both teacher and student probabilities in our framework. Recall that $\partial \mathcal{L}_{\mathrm{SR}} / \partial \tilde{b}^{l}_x$ is propagated to  $b^l_x$ identically in Eq.~\ref{eq:perturb}. Each model parameter is therefore updated based on the joint effects from $\bm{p}_x$ and $\bm{\tilde{p}}_x$. This avoids excessively large gradients from any single probability. We observe degraded performance when either the gradients of teacher probabilities in KL divergence or that of $b^l_x$ are blocked.

\begin{algorithm}[t]
	\renewcommand{\algorithmicrequire}{\textbf{Input:}}
	\renewcommand{\algorithmicensure}{\textbf{Initialization:}}
	\caption{Safe Training Mechanism.} 
	\label{alg:st} 
	\begin{algorithmic}[1]
		%	\Require Model $\mathcal{M}$, $div$, $T_r$, $t_r$ and checking level $L$, current step $iter$.
		\Ensure $last\_restore=0$,  save snapshot of $\mathcal{M}$
		\Procedure{CheckDivDrop}{$\mathrm{div}$, $L$, $T$, $iter$}
		\For{$l=1$ \textbf{to} $L$} \gray{\Comment{check at multi-scales}}
		\State $divs=\mathrm{div}(iter-T, \dots, iter)$ \gray{\Comment{get diversity}}
		\State $divs = \mathrm{split}(divs, 2^l)$ \gray{\Comment{to even sub-intervals}}
		\For{$i=0$ \textbf{to} $\mathrm{len}(divs)-1$}
		\If{$\mathrm{avg}(divs[i+1])<\mathrm{avg}(divs[i])-1$}
		\State \textbf{return} True \gray{\Comment{significant dropping}}
		\EndIf
		\EndFor
		\EndFor
		\State \textbf{return} False
		\EndProcedure
		\State				
		\Procedure{SafeTraining}{$\mathcal{M}$, $\textrm{div}$, $T$, $L$, $iter$}
		\If{$iter\ \%\ T==0$ \textbf{and} $iter>=T$}
		\If{\textsc{CheckDivDrop}($\mathrm{div}$, $L$, $T$, $iter$)}
		\State Restore $\mathcal{M}$ to last snapshot, $t_r=iter$
		\If {$iter-last\_restore\leq T_r$}		
		\State $T_r = T_r \times 2$ \gray{\Comment{avoid oscillation}}
		\EndIf 
		\State $last\_restore=iter$ 
		\EndIf 
		\State Save snapshot of $\mathcal{M}$
		\EndIf	
		\State \Return $\mathcal{M}$, $T_r$, $t_r$
		\EndProcedure		
	\end{algorithmic}
\end{algorithm}

\begin{algorithm}[t]
	\renewcommand{\algorithmicrequire}{\textbf{Input:}}
	\renewcommand{\algorithmicensure}{\textbf{Initialization:}}
	\caption{SSRT algorithm.} 
	\label{alg:ssd} 
	\begin{algorithmic}[1]
		\Require Model $\mathcal{M}$, source data $\mathcal{D}_s$, target data $\mathcal{D}_t$, confidence threshold $\epsilon$, self-refinement loss weight $\beta$, perturbation scalar $\alpha$, Safe Training parameters $T$ and $L$, diversity measure $\textrm{div}(\cdot)$.
		\Ensure $T_r=T$, $t_r=0$
		\For{$iter=0$ \textbf{to} $max\_Iter$}
		\State Sample a batch from source data and target data
		\State Obtain $r$ via Eq.~\ref{eq:r}, $\alpha_{r}=r\alpha$, $\beta_{r}=r\beta$         
		\State Randomly choose $l\in\{0,4,8\}$, add perturbation via Eq.~\ref{eq:perturb} using $\alpha_{r}$, obtain $\mathcal{L}_{\mathrm{SR}}$ via Eq.~\ref{eq:SD}
		\State Update model parameters via Eq.~\ref{eq:opt} using $\beta_{r}$
		\State $\mathcal{M},T_r,t_r \gets\textsc{SafeTraining}(\mathcal{M},\textrm{div},T,L,iter)$
		\EndFor
	\end{algorithmic}
\end{algorithm}

\subsection{Safe Training via Adaptive Adjustment}

In the proposed self-refinement strategy, setting a proper value of the perturbation scalar $\alpha$ and the self-refinement loss weight $\beta$ is critical. Excessively large perturbations lead to a collapse of the predicted class distribution, while a small one may under-exploit its benefit. Since the target domain is fully unlabeled and domain adaptation tasks vary widely even for the same dataset, it is desired to adjust these values adaptively. Some works~\cite{tarvainen2017mean, berthelot2019mixmatch} apply a ramp-up period at the beginning of training. While this alleviates the tendency to collapse during this period, it cannot solve the issue when the maximum value is improper for current adaptation tasks.

\begin{table*}[!t]
	\caption{Accuracies (\%) on \textbf{Office-Home}. $^*$CDTrans uses DeiT-base backbone. $^\circ$TVT uses ViT-base backbone. ``-S/B" indicates ViT-small/base backbones, respectively. }
	\footnotesize
	\centering
	\scalebox{0.9}{
	\begin{tabular}{p{1.8cm}p{0.74cm}<{\centering}p{0.74cm}<{\centering}p{0.74cm}<{\centering}p{0.74cm}<{\centering}p{0.74cm}<{\centering}p{0.74cm}<{\centering}p{0.74cm}<{\centering}p{0.74cm}<{\centering}p{0.74cm}<{\centering}p{0.74cm}<{\centering}p{0.74cm}<{\centering}p{0.74cm}<{\centering}>{\columncolor{tbgray}}p{0.74cm}<{\centering}}
		\toprule
		Method & Ar$\shortrightarrow$Cl & Ar$\shortrightarrow$Pr & Ar$\shortrightarrow$Rw &  Cl$\shortrightarrow$Ar & Cl$\shortrightarrow$Pr & Cl$\shortrightarrow$Rw &  Pr$\shortrightarrow$Ar & Pr$\shortrightarrow$Cl & Pr$\shortrightarrow$Rw &      
		Rw$\shortrightarrow$Ar & Rw$\shortrightarrow$Cl & Rw$\shortrightarrow$Pr & Avg.   \\ 	
		\midrule
		ResNet-50~\cite{he2016deep} & 34.9 & 50.0 & 58.0 & 37.4 & 41.9 & 46.2 & 38.5 & 31.2 & 60.4 & 53.9 & 41.2 & 59.9 & 46.1\\
		CDAN+E~\cite{long2018conditional} & 50.7 & 70.6 & 76.0 & 57.6 & 70.0 & 70.0 & 57.4 & 50.9 & 77.3 & 70.9 & 56.7 & 81.6 & 65.8\\
		SAFN~\cite{xu2019larger} & 52.0 & 71.7 & 76.3 & 64.2 & 69.9 & 71.9 & 63.7 & 51.4 & 77.1 & 70.9 & 57.1 & 81.5 & 67.3 \\
		CDAN+TN~\cite{wang2019transferable} & 50.2 & 71.4 & 77.4 & 59.3 & 72.7 & 73.1 & 61.0 & 53.1 & 79.5 & 71.9 & 59.0 & 82.9 & 67.6\\
		SHOT~\cite{liang2020we} & 57.1 & 78.1 & 81.5 & 68.0 & 78.2 & 78.1 & 67.4 & 54.9 & 82.2 & 73.3 & 58.8 & 84.3 & 71.8\\
		DCAN+SCDA~\cite{li2021semantic} & 60.7 & 76.4 & 82.8 & 69.8 & 77.5 & 78.4 & 68.9 & 59.0 & 82.7 & 74.9 & 61.8 & 84.5 & 73.1\\			
		 CDTrans$^*$\cite{xu2021cdtrans} & 68.8 & 85.0 & 86.9 & 81.5 & 87.1 & 87.3 & 79.6 & 63.3 & 88.2 & 82.0 & 66.0 & 90.6 & 80.5 \\
		TVT$^\circ$ \cite{yang2021tvt} & 74.89 & 86.82 & 89.47 & 82.78 & 87.95 & 88.27 & 79.81 & 71.94 & 90.13 & 85.46 & 74.62 & 90.56  &83.56 \\
		% CDTrans\cite{xu2021cdtrans} & 66.1 & 88.9 & 89.5 & 83.8 & \HL{88.8} & 89.1 & 81.5 & 61.9 & 90.5 & 83.2 & 62.3 & 90.6 & 81.3\\
		\midrule
		ViT-S~\cite{dosovitskiy2020image} & 47.01 &	76.98 &	83.54 &	69.84 & 77.11 &	80.42 &	68.15 &	44.08 &	82.86 &	74.78 &	47.97 &	84.66 &	69.78 \\
		Baseline-S & 59.59 & 80.11 &	84.67 &	73.84 &	78.49 &	81.36 &	74.41 &	59.82 &	86.27 &	80.10 &	62.59 &	87.23 & 75.71 \\
		SSRT-S (ours) & 67.03 &	84.21 &	88.32 &	79.85 &	84.28 &	87.58 &	80.72 &	66.03 &	88.27 &	82.04 &	69.44 &	89.86 &	80.64 \\
		ViT-B~\cite{dosovitskiy2020image} & 54.68 &	83.04 &	87.15 &	77.30 &	83.42 &	85.54 &	74.41 &	50.90 &	87.22 &	79.56 &	53.79 &	88.80 &	75.48 \\
		Baseline-B & 66.96 &	85.74 &	88.07 &	80.06 &	84.12 &	86.67 &	79.52 &	67.03 &	89.44 &	83.64 &	70.15 &	91.17 &	81.05 \\
		Baseline-B+MI & 70.63 &	88.62 &	89.99 &	82.08 &	87.84 &	89.28 &	81.01 &	68.82 &	\HL{91.26} &	85.17 &	71.66 &	\HL{92.45} &	83.23\\
		%Mixup & 71.32 &	86.66 &	88.82 &	82.45 &	84.79 &	87.58 &	82.90 &	71.68 &	90.77 &	85.46 &	74.36 &	91.37 &	83.18\\
		%VAT & 74.32 &	87.05 &	89.72 &	83.68 &	87.59 &	89.21 &	82.41 &	71.39 &	89.99 &	85.58 &	75.67 &	92.52 &	84.09\\
		%Ours (self-aug) & 73.49 &	88.26 &	91.16 &	85.00 &	87.68 &	89.56 &	85.37 &	74.39 &	91.16 &	86.07 &	76.22 &	92.59 &	85.08 \\			
		%Ours (no period) & 74.71 &	88.69 & 	91.26 & 	85.21 &	87.88 &	89.58 &	85.74 &	74.98 &	91.16 &	85.46 &	78.03 &	91.98 &	85.39 \\
		SSRT-B (ours) & \HL{75.17} &	\HL{88.98} &	\HL{91.09} &	\HL{85.13} &	\HL{88.29} &	\HL{89.95} &	\HL{85.04} &	\HL{74.23} &	\HL{91.26} &	\HL{85.70} &	\HL{78.58} &	91.78 &	\HL{85.43} \\
		\bottomrule
	\end{tabular} }
	\label{tab:officehome}
\end{table*}

\begin{figure*}[h]
	\centering
	\includegraphics[width=0.9\linewidth]{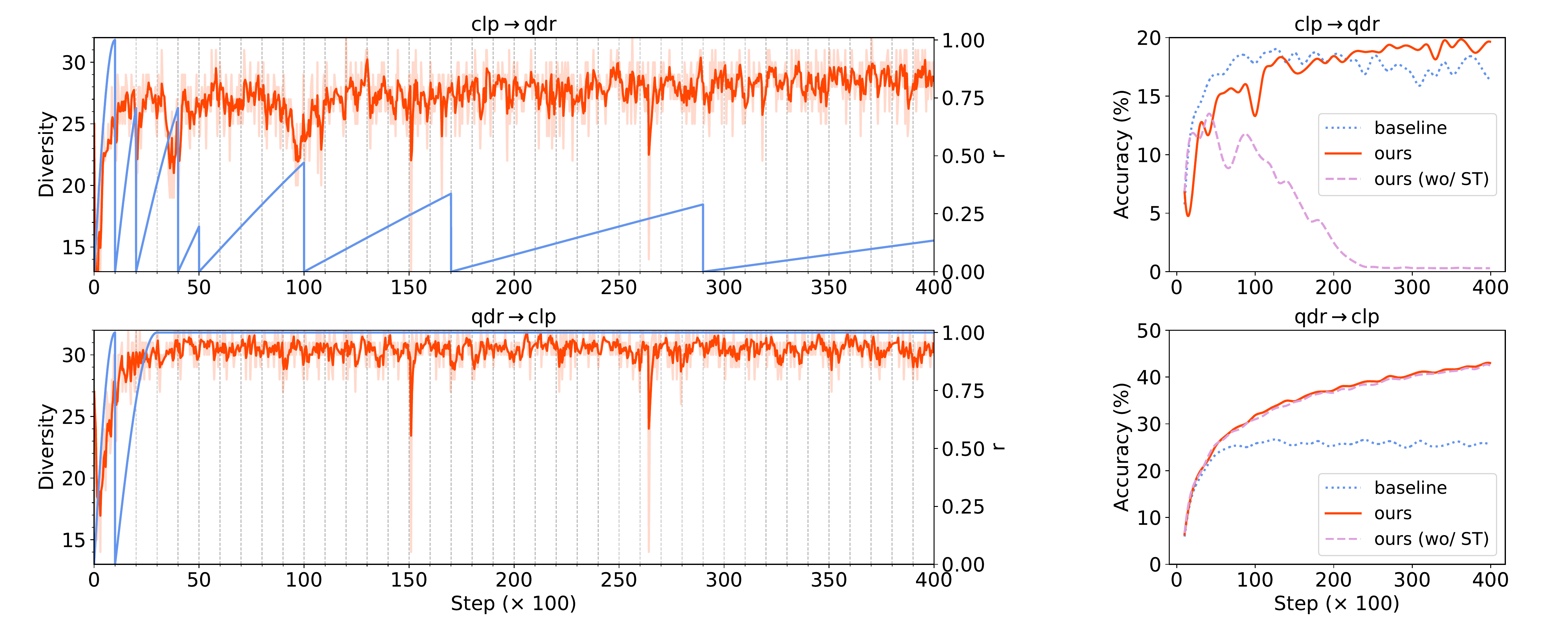}
	\caption{Representative training histories using Safe Training (ST) on DomainNet clp$\rightarrow$qdr and qdr$\rightarrow$clp. (\textbf{Left}) Plots of the diversity of model predictions on target domain data and the adaptive scalar $r$. For better visualization, both original values (light color) and smoothed values (dark color) of diversity are shown. (\textbf{Right}) Plots of comparison test accuracies on target domain data.}
	\label{fig:period}
\end{figure*}

\begin{table*}[!t]
	\caption{Accuracies (\%) on \textbf{DomainNet}. In each sub-table, the column-wise means source domain and the row-wise means target domain.}
	\scriptsize
	\centering	
	\renewcommand{\arraystretch}{1.}
	\scalebox{0.98}{
	\input{Table_DomainNet_deit.latex}
	}
	\label{tab:domainnet}
\end{table*}

\begin{table*}[!t]
	\caption{Accuracies (\%) on \textbf{VisDA-2017}.}
	\footnotesize
	\centering
	\scalebox{0.9}{
	\begin{tabular}{p{1.9cm}p{0.74cm}<{\centering}p{0.74cm}<{\centering}p{0.74cm}<{\centering}p{0.74cm}<{\centering}p{0.74cm}<{\centering}p{0.74cm}<{\centering}p{0.74cm}<{\centering}p{0.74cm}<{\centering}p{0.74cm}<{\centering}p{0.74cm}<{\centering}p{0.74cm}<{\centering}p{0.74cm}<{\centering}>{\columncolor{tbgray}}p{0.74cm}<{\centering}}
		\toprule
		Method  &  plane & bcycl & bus & car & horse & knife & mcycl & person & plant & sktbrd & train & truck & Avg.  \\ \midrule
		ResNet-101~\cite{he2016deep}  & 55.1  & 53.3  & 61.9  & 59.1  & 80.6  & 17.9  & 79.7  & 31.2  & 81.0  & 26.5  & 73.5  & 8.5  & 52.4 \\
		DANN~\cite{ganin2015unsupervised}  & 81.9  & 77.7  & 82.8  & 44.3  & 81.2  & 29.5  & 65.1  & 28.6  & 51.9  & 54.6  & 82.8  & 7.8  & 57.4 \\
		CDAN~\cite{long2018conditional} & 85.2 & 66.9 & 83.0 & 50.8 & 84.2 & 74.9 & 88.1 & 74.5 & 83.4 & 76.0 & 81.9 & 38.0 & 73.9 \\
		SAFN~\cite{xu2019larger}  & 93.6  & 61.3  & 84.1  & 70.6  & 94.1  & 79.0  & 91.8  & 79.6  & 89.9  & 55.6  & 89.0  & 24.4  & 76.1\\
		SWD~\cite{lee2019sliced} & 90.8 & 82.5 & 81.7 & 70.5 & 91.7 & 69.5 & 86.3 & 77.5 & 87.4 & 63.6 & 85.6 & 29.2 & 76.4 \\
		SHOT~\cite{liang2020we} & 94.3 & 88.5 & 80.1 & 57.3 & 93.1 & 94.9 & 80.7 & 80.3 & 91.5 & 89.1 & 86.3 & 58.2 & 82.9 \\
		% CDtrans\cite{xu2021cdtrans} & 97.2 & \HL{88.4} & 86.5 & 81.2 & \HL{98.7} & 95.4 & 94.1 & 80.0 & \HL{95.2} & 97.5 & 94.4 & 48.4 & 88.1\\
		CDTrans$^*$\cite{xu2021cdtrans} & 97.1 & 90.5 & 82.4 & 77.5 & 96.6 & 96.1 & 93.6 & \HL{88.6} & \HL{97.9} & 86.9 & 90.3 & \HL{62.8} & 88.4 \\
		TVT$^\circ$ \cite{yang2021tvt} & 92.92 & 85.58 & 77.51 & 60.48 & 93.60 & 98.17 & 89.35 & 76.40 & 93.56 & 92.02 & 91.69  & {55.73} & 83.92\\
		\midrule
		ViT-B~\cite{dosovitskiy2020image} & 99.09 &	60.66 &	70.55 &	82.66 &	96.50 &	73.06 &	\HL{97.14} &	19.73 &	64.48 &	94.74 &	97.21 &	15.36 &	72.60 \\
		Baseline-B & 98.55 &	82.59 &	85.97 &	57.07 &	94.93 &	97.20 &	94.58 &	76.68 &	92.11 &	96.54 &	94.31 &	52.24 &	85.23 \\
		Baseline-B+MI & 98.63 &	\HL{90.79} &	81.83 &	47.28 &	96.29 &	98.36 &	84.68 &	70.70 &	93.30 &	97.54 &	\HL{94.55} &	55.03 &	84.08 \\
		%Mixup &  & & & & & & & & & & & & 88.21 \\
		%VAT &  & & & & & & & & & & & & 89.24 \\
		%\HL{FixMatch} & 87.98  & & & & & & & & & & & & 89.93\\
		%\midrule
		%Ours (self-aug) & & & & & & & & & & & & & 88.25 \\
		%Ours (no period) & 98.99 &	88.58 &	87.59 &	84.17 &	98.36 &	98.94 &	95.39 &	85.95 &	94.09 &	97.28 &	94.43 &	41.33 &	88.76 \\	
		SSRT-B (ours) & \HL{98.93} &	87.60 &	\HL{89.10} &	\HL{84.77} &	\HL{98.34} &	\HL{98.70} &	96.27 &	{81.08} &	94.86 &	\HL{97.90} &	94.50 &	43.13 &	\HL{88.76} \\	
		\bottomrule
	\end{tabular}  	}
	\label{tab:visda}
	\vspace{-2mm}
\end{table*}

We propose a \emph{Safe Training} mechanism. The observation is that whenever the model begins to collapse, the diversity of model predictions will decrease simultaneously. Our goal is to detect such events while monitoring the training process. Once it occurs, the learning configuration is reset and meanwhile the model is restored to a previously achieved state. Specifically, an adaptive scalar $r\in [0,1]$ is adopted to modulate $\alpha$ and $\beta$, \ie, $\alpha_{r}=r\alpha$ and $\beta_{r}=r\beta$. We define a fixed period $T$ and divide the training process into consecutive intervals. A model snapshot is saved at the end of each interval. Then $r$ is defined as
\begin{equation}
	r(t)=
	\begin{cases}	
	\sin\left(\frac{\pi}{2T_r}(t-t_r)\right) & {\rm if}\ t-t_r < T_r\\
	1.0 & {\rm otherwise}
	\end{cases},
	\label{eq:r}
\end{equation}
where $t$ is the current training step. Initially, $T_r=T$ and $t_r=0$. It hence takes $T$ steps for $r$ to ramp up to 1.0. At the end of each interval, the diversity of model predictions within this interval is checked to find abrupt dropping. If not existed, the formulation of $r$ remains unchanged. Otherwise, $t_r$ is reset to current training step $t$, and the model is restored to last snapshot. To avoid oscillation between collapse and restoration, $T_r$ is doubled if the last restoration occurs within $T_r$ steps. Figure~\ref{fig:SD} illustrates the training process with adaptive adjustment. Two events of diversity dropping are identified (marked with pink areas), leading to two model restorations and reset of $r$.

The remaining issue is which diversity measure to use and how to detect diversity dropping. We find that the number of unique model predicted labels on each target training batch $\mathcal{B}_t$ works well. We hence define the following diversity measure: 
\begin{equation}
	\textrm{div}(t;\mathcal{B}_t) = unique\_labels(h(\mathcal{B}_t)).
\end{equation}
To detect diversity dropping, we split the interval into sub-intervals and check whether the average diversity value drops across each sub-interval. We implement this at multi-scales to improve the sensitivity of detection. Every consecutive sub-intervals of $T/2^1, \cdots, T/2^L$ steps are checked for a given integer $L$.  Details are listed in Alg.~\ref{alg:st} and Alg.~\ref{alg:ssd}.

\section{Experiments}
\label{sec:exp}

We evaluate our method on four popular UDA benchmarks. \textbf{Office-31}~\cite{saenko2010adapting} contains 4,652 images of 31 classes from three domains: Amazon (A), DSLR (D) and Webcam (W). \textbf{Office-Home}~\cite{venkateswara2017deep} consists of 15,500 images of 65 classes from four domains: Artistic (Ar), Clip Art (Cl), Product (Pr), and Real-world (Rw) images.  \textbf{VisDA-2017}~\cite{peng2017visda} is a Synthetic-to-Real dataset, with about 0.2 million images in 12 classes. \textbf{DomainNet}~\cite{peng2019moment} is the largest DA dataset containing about 0.6 million images of 345 classes in 6 domains: Clipart (clp), Infograph (inf), Painting (pnt), Quickdraw (qdr), Real (rel), Sketch (skt).

We use the ViT-base and ViT-small with 16$\times$16 patch size ~\cite{dosovitskiy2020image, steiner2021augreg}, pre-trained on ImageNet~\cite{ILSVRC15}, as the vision transformer backbones. For all tasks, we use an identical set of hyper-parameters ($\alpha=0.3$, $\beta=0.2$, $\epsilon=0.4$, $T=1000$, $L=4$). Ablation studies on them are provided in Sec.~\ref{sec:abl}. More details can be found in the supplementary material.

Our comparison methods include DANN~\cite{ganin2015unsupervised}, CDAN~\cite{long2018conditional}, CDAN+E~\cite{long2018conditional}, SAFN~\cite{xu2019larger}, SAFN+ENT~\cite{xu2019larger}, CDAN+TN~\cite{wang2019transferable}, SHOT~\cite{liang2020we}, DCAN+SCDA~\cite{li2021semantic}, MDD+SCDA~\cite{li2021semantic}, SWD~\cite{lee2019sliced}, MIMTFEL~\cite{gao2020reducing}, TVT~\cite{yang2021tvt} and CDTrans~\cite{xu2021cdtrans}. ``Baseline" is ViT with adversarial adaptation (see Sec.~\ref{sec:background}). We also include its combination with Mutual Information (MI) loss~\cite{gomes2010discriminative,shi2012information} in comparison.

\subsection{Results on Benchmarks}
Tables~\ref{tab:officehome}-\ref{tab:office31} present evaluation results on four benchmarks. We use ``-S/B” to indicate results using ViT-small/base backbones, respectively. For Office-Home and Offce-31, CNN-based methods use ResNet-50 as their backbones; whereas for DomainNet and VisDA they use ResNet-101. Generally, the transformer-based results are much better. This is attributed to its strong transferable feature representations. ViT-base is better than ViT-small, due to higher model complexity. Apparently, Baselines improve over source-only training. Integrating Mutual Information (MI) loss further improves. Compared with other methods, SSRT-B performs the best on Office-Home, DomainNet and VisDA. It improves 4.38\% on Office-Home, 3.53\% on VisDA-2017 and 6.7\% on DomainNet over Baseline-B despite that Baseline-B is already very strong. In particular, on the challenging DomainNet dataset, SSRT-B achieves an impressive 45.2\% average accuracy. It is worth mentioning that in DomainNet some domains have large gaps from the others, such as \emph{inf} and \emph{qdr}. Transferring among these domains and others is very difficult. It is thus desired to transfer safely and not deteriorate the performance significantly. Looking at tasks with \emph{qdr} being target domain, SSRT-B obtains 29.3\% average accuracy, while many other methods perform poorly. We illustrate the effects of some important components that contribute to our excellent performance in the following sections.  

 \begin{table}[t]
	\caption{Accuracies (\%) on \textbf{Office-31}.} 
	\centering
	\footnotesize
	%\begin{tabular}{lccccccc}
	%\begin{tabular}{p{1.8cm}p{1.5cm}<{\centering}p{1.5cm}<{\centering}p{1.5cm}<{\centering}p{1.5cm}<{\centering}p{1.5cm}<{\centering}p{1.5cm}<{\centering}p{1.5cm}<{\centering}}
	\begin{tabular}{@{}p{1.8cm}p{0.48cm}<{\centering}p{0.48cm}<{\centering}p{0.48cm}<{\centering}p{0.48cm}<{\centering}p{0.48cm}<{\centering}p{0.48cm}<{\centering}>{\columncolor{tbgray}}p{0.48cm}<{\centering}}
		
		\toprule
		Method                        & A$\shortrightarrow$W & D$\shortrightarrow$W & W$\shortrightarrow$D & A$\shortrightarrow$D & D$\shortrightarrow$A & W$\shortrightarrow$A & Avg. \\ 
		\midrule	
		ResNet-50~\cite{he2016deep} & 68.4 & 96.7 & 99.3 & 68.9 & 62.5 & 60.7 & 76.1\\
		DANN~\cite{ganin2015unsupervised} & 82.0 & 96.9 & 99.1 & 79.7 & 68.2 & 67.4 & 82.2 \\			
		SAFN+ENT~\cite{xu2019larger} & 90.1 & 98.6 & 99.8 & 90.7 & 73.0 & 70.2 & 87.1 \\
		CDAN+TN~\cite{wang2019transferable} & 95.7 & 98.7 &  100. & 94.0 & 73.4 & 74.2 & 89.3\\
		SHOT~\cite{liang2020we} & 90.1 & 98.4 & 99.9 & 94.0 & 74.7 & 74.3 & 88.6\\
		MDD+SCDA~\cite{li2021semantic} & 95.3 & 99.0 & 100. & 95.4 & 77.2 & 75.9 & 90.5 \\
		CDTrans$^*$\cite{xu2021cdtrans} & 96.7 & 99.0 & 100. & 97.0 & 81.1 & 81.9 & 92.6 \\
		TVT$^\circ$\cite{yang2021tvt} & 96.4 & {99.4} & {100.} & 96.4 & {84.9} & {86.1} & {93.8}\\
		\midrule
		ViT-S~\cite{dosovitskiy2020image} & 86.9 &	98.6 &	100.  &	88.6  &	76.0  &	75.9  &	87.7 \\
		Baseline-S &  91.9 &	99.1 &	100.  &	89.2 &	78.4 &	77.9 &	89.4 \\
		SSRT-S (ours) & 95.7 &	99.2 &	100.  &	95.8  &	79.2  &	79.9  &	91.6 \\
		ViT-B~\cite{dosovitskiy2020image} & 91.2 &	99.2 &	{100.} &	90.4 &	81.1 &	80.6 &	90.4 \\	
		Baseline-B & 92.5 &	99.2 &	{100.} &	93.6 &	80.7 &	80.7 &	91.1 \\
		%Ours (no period) &  97.99 &	99.25 &	100.0 & 97.19 & 82.78 &	80.80 &	93.00 \\
		SSRT-B (ours) & {97.7} &	99.2 &	{100.} &	{98.6} &	83.5 &	82.2 &	93.5
		\\			
		\bottomrule
	\end{tabular}
	\label{tab:office31}
	\vspace{-2mm}
\end{table}

\begin{table}[!t]
	\caption{Accuracies (\%) compared with perturbing raw inputs. $X^\dagger$ means averaged over all 5 tasks with $X$ being the target domain.} 
	\centering
	\footnotesize
	\renewcommand\arraystretch{1.2}
	\begin{tabular}{@{}p{1.6cm}@{\hspace{3mm}}>{\columncolor{tbgray}}p{0.55cm}<{\centering}>{\columncolor{tbgray}}p{0.78cm}<{\centering}p{0.32cm}<{\centering}p{0.32cm}<{\centering}p{0.32cm}<{\centering}p{0.32cm}<{\centering}p{0.32cm}<{\centering}p{0.32cm}<{\centering}}	
		\toprule
		& OH & DN &  clp$^\dagger$ &	inf$^\dagger$ & pnt$^\dagger$  &	qdr$^\dagger$  &	rel$^\dagger$	 & skt$^\dagger$ \\
		\midrule
		Baseline-B & 81.1 & 38.5 &
		50.6 &	25.6 &	44.9 &	11.6 &	57.0 &	41.5
		\\ 		
		SSRT-B (raw) & 85.0 & 44.2 & 58.6 &	26.7 &	51.7 &	13.7 &	63.9 &	50.8 \\		
		SSRT-B & 85.4 & 45.2 & 60.0 &	28.2 &	53.3 &	13.7 &	65.3 &	50.4 \\		
		\bottomrule
	\end{tabular}
	\label{tab:layer}
	\vspace{-2mm}
\end{table}

\subsection{Effects of Multi-layer Perturbation}

Table~\ref{tab:layer} verifies that applying perturbation to the latent token sequences performs better than to the raw input images on Office-Home (OH) and DomainNet (DN). Fig.~\ref{fig:abl:layer} compares performances when adding the same amount of perturbation to each layer while not using safe training. As can be seen, the best layer to apply perturbation varies across tasks. Besides, a layer that works for one task may fail on others. In our experiments, we uniformly choose one layer from \{0,4,8\}. As a comparison, perturbing any single layer from it decreases the average accuracy on DomainNet by -1.0\%, -1.5\% and -1.5\%, respectively.

\begin{figure}[t]
	\centering
	\begin{subfigure}{0.49\linewidth}
		\centering
		\includegraphics[width=1.0\linewidth]{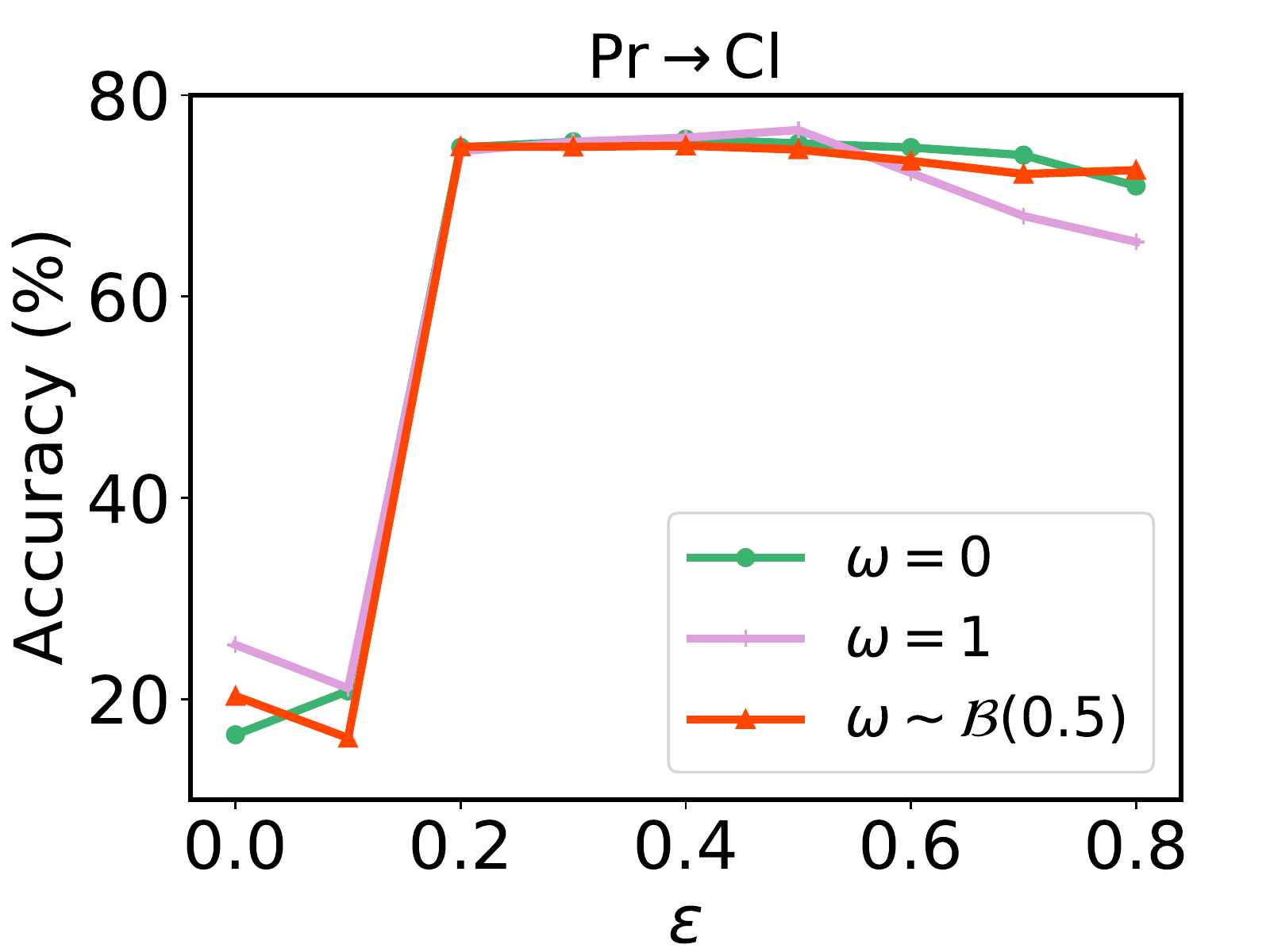}
		%\caption{}
	\end{subfigure}
	%\hfill
	\begin{subfigure}{0.49\linewidth}
		\centering
		\includegraphics[width=1.0\linewidth]{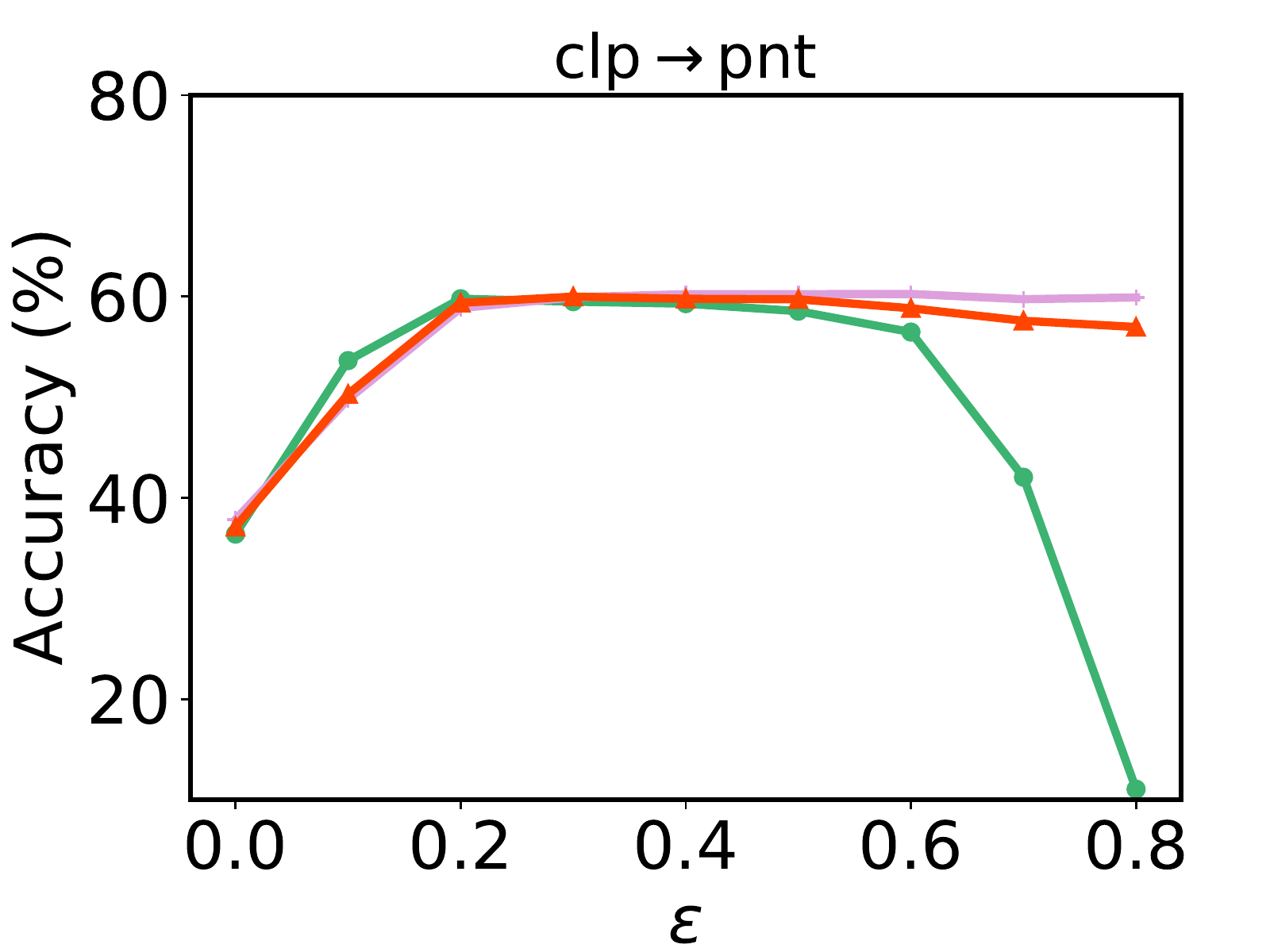}
		%\caption{}
	\end{subfigure}
	\begin{subfigure}{0.49\linewidth}
		\centering
		\includegraphics[width=1.0\linewidth]{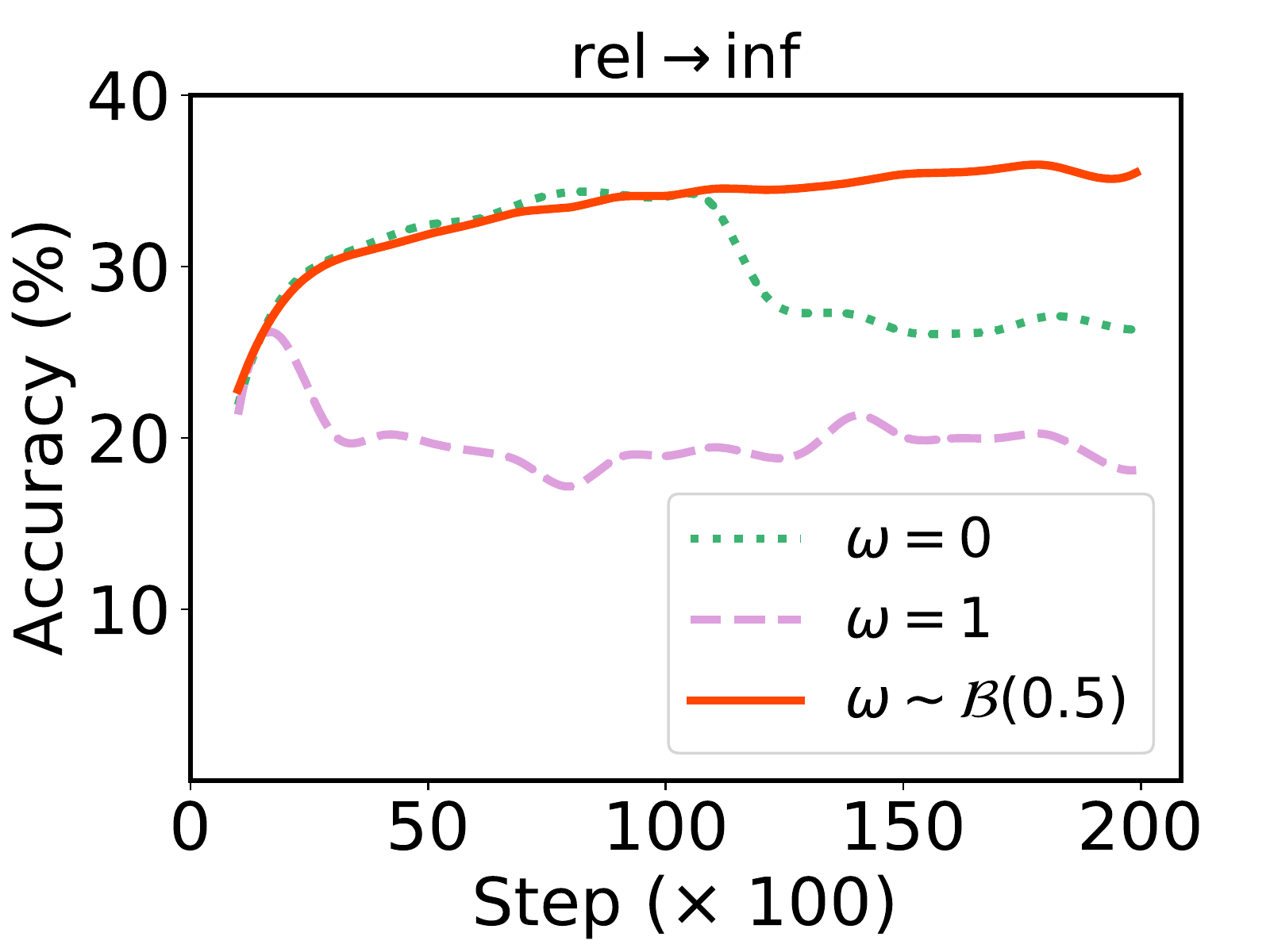}
		%\caption{}
	\end{subfigure}
	%\hfill
	\begin{subfigure}{0.49\linewidth}
		\centering
		\includegraphics[width=1.0\linewidth]{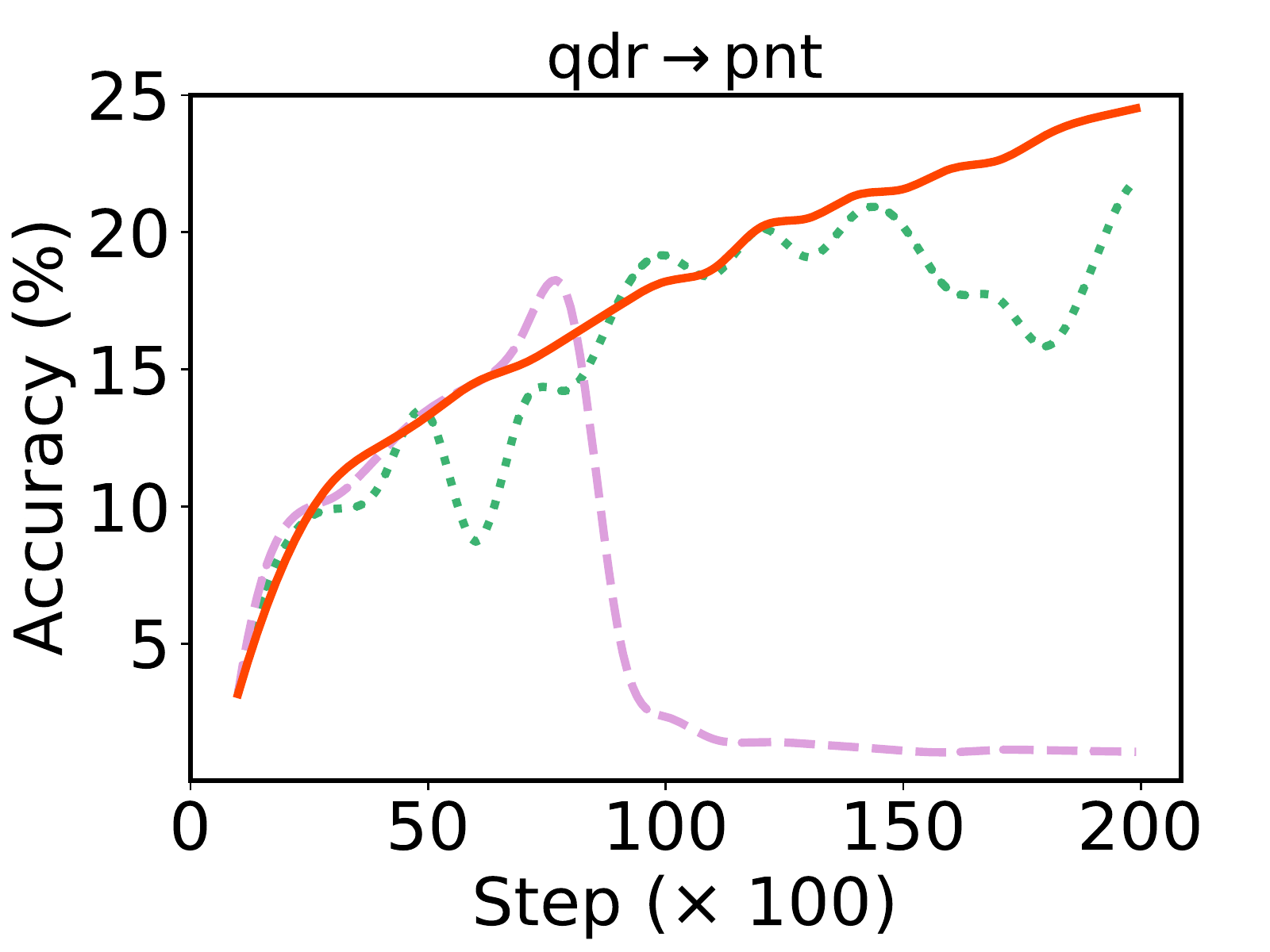}
		%\caption{}
	\end{subfigure}
	\caption{Comparison of self-refinement losses. (\textbf{Upper}) Varying confidence threshold $\epsilon$. (\textbf{Lower}) Test accuracies on target domain data. (Safe Training not applied)}
	\label{fig:loss_func}
\end{figure}

\begin{table}[t]
	\caption{Accuracies (\%) using comparison losses. All results are reported at training step of 20k. $X^\dagger$ means averaged over all 5 tasks with $X$ being target domain. $^\ddagger$ Using Safe Training. } 
	\centering
	\footnotesize
	\renewcommand\arraystretch{1.2}
	\begin{tabular}{@{}p{1.5cm}>{\columncolor{tbgray}}p{0.56cm}<{\centering}>{\columncolor{tbgray}}p{0.78cm}<{\centering}p{0.32cm}<{\centering}p{0.32cm}<{\centering}p{0.32cm}<{\centering}p{0.32cm}<{\centering}p{0.32cm}<{\centering}p{0.32cm}<{\centering}}	
		\toprule
		& OH & DN &  clp$^\dagger$ &	inf$^\dagger$ & pnt$^\dagger$  &	qdr$^\dagger$  &	rel$^\dagger$	 & skt$^\dagger$ \\
		\midrule
		Baseline-B & 81.1 & 38.9 & 50.7 & 25.5 &	46.1 & 11.9 & 57.4 & 42.0\\ 
		$\omega=0$ & 85.5 & 41.1 & 57.3 & 22.0 & 52.2 & 1.8 & 63.4 &	49.9  \\
		$\omega=1$ & 85.7 & 40.1 & 56.6 & 23.4 &	48.1 &	0.3 & 63.3 & 49.0  \\
		$\omega\!\sim\!\mathcal{B}(0.5)$ & 85.4 & 41.8 & 57.0 & 26.6 &	53.0 & 1.8 & 63.2 &	49.5  \\	
		$\omega\!\sim\!\mathcal{B}(0.5)^\ddagger$ & 85.4 & 43.4 & 57.0 &	28.2 &	51.8 &	13.0 &	62.9 &	47.4  \\		
		\bottomrule
	\end{tabular}
	\label{tab:all_loss_func}
	\vspace{-4mm}
\end{table}

\subsection{Effects of Bi-directional Self-Refinement}
\label{sec:bisup}

Our method adopts bi-directional supervision for self-refinement in Eq.~\ref{eq:SD}. The main consideration is to improve method's safeness. Figure~\ref{fig:loss_func} compares with uni-directional self-refinement by fixing $\omega$ to be 0 or 1. In the upper two figures, their performance drops for relatively large confidence threshold $\epsilon$. In the lower two figures, model collapse occurs after training for some steps. In contrast, bi-directional self-refinement is more robust as it combines the two losses, thus reduces the negative effect of either one. Table~\ref{tab:all_loss_func} presents some quantitative results. On Office-Home, all losses perform similarly well. On  DomainNet, bi-directional self-refinement works better. However, they all fail on challenging tasks when target domain is \emph{qdr}. This is solved with Safe Training.

Another important issue is when to back-propagate gradients. Table ~\ref{tab:detach} shows that the performance degrades when either the gradient for $b^l_x$ in Eq.~\ref{eq:perturb} or the teacher probability of KL divergence in Eq.~\ref{eq:SD} are blocked. An interesting finding is that the bi-directional self-refinement appears to be more robust even when the gradients are blocked. We believe this is because the two losses are complementary.

\begin{table}[t]
	\caption{Blocking gradient back-propagation for different variables. Note that $\bm{p}_x$ and $\bm{\tilde{p}}_x$ in the table only refer to the teacher probability in KL divergence. (Safe Training not applied)} 
	\centering
	\footnotesize
	\begin{tabular}{p{1.6cm}|p{0.4cm}<{\centering}p{0.4cm}<{\centering}p{0.4cm}<{\centering}|p{0.8cm}<{\centering}p{0.8cm}<{\centering}p{0.8cm}<{\centering}}	
	 \toprule
	  & $b^l_x$ & $\bm{p}_x$ & $\bm{\tilde{p}}_x$  &Pr$\shortrightarrow$Ar & Pr$\shortrightarrow$Cl &  Pr$\shortrightarrow$Rw \\ % & Cl$\shortrightarrow$Ar &Cl$\shortrightarrow$Pr &  Cl$\shortrightarrow$Rw 
	 \midrule
	 $\omega=0$ &  &   & $\times$ & 4.70 &2.66 & 16.39\\%   &	1.61 & 12.71 &	6.08 \\
	 $\omega=1$ &  & $\times$ &  & 79.15 &44.38	 & 89.14 \\%  &	81.17 & 85.00 &	87.28  \\
	 $\omega\sim\mathcal{B}(0.5)$ &  & $\times$ & $\times$ & 84.10 & 71.32  & 90.75\\%   & 83.68 & 85.69 &	88.04 \\
	 \midrule
	 $\omega\sim\mathcal{B}(0.5)$ & $\times$ &  &  & 84.38 & 72.60  & 90.87\\%  &	84.55 & 87.27 & 89.49 \\
	 $\omega\sim\mathcal{B}(0.5)$ &  &  &  & 85.74 & 74.98  & 91.16  \\% &	85.21 & 87.88 &	89.58 \\
	 \bottomrule
\end{tabular}
\label{tab:detach}
\end{table}

\subsection{Effects of Safe Training}
As observed previously, the vanilla training strategy may fail on some tasks. The reason is that the predicted class distribution on target domain data collapses due to excessive perturbation or too large loss weight, even if they work well on other tasks. Safe Training adjusts their values adaptively to avoid such situation. Figure~\ref{fig:period} presents detailed training histories on two representative tasks to show how it works. For \emph{qdr$\rightarrow$clp}, the adaptive scalar $r$ quickly converges to 1.0 and the diversity stabilizes to a relatively high value. Training model with or without Safe Training performs similarly. For \emph{clp$\rightarrow$qdr}, diversity drops after some steps, and $r$ resets to smaller values. A clear correlation between diversity and accuracy can be observed. For example, at step of 10k, the accuracy drops abruptly and diversity drops concurrently. Without Safe Training, model collapses after about 10k iterations. With Safe Training, the model trains normally and surpasses the baseline finally. It should be noted that model collapse mainly affects target domain data. For \emph{clp$\rightarrow$qdr} without safe training, the final accuracy on source domain is 96.9\% while that on target domain is only 0.3\%.

\subsection{Visualization of Perturbation}

\begin{figure}[!t]
	\centering
	\includegraphics[width=1.0\linewidth]{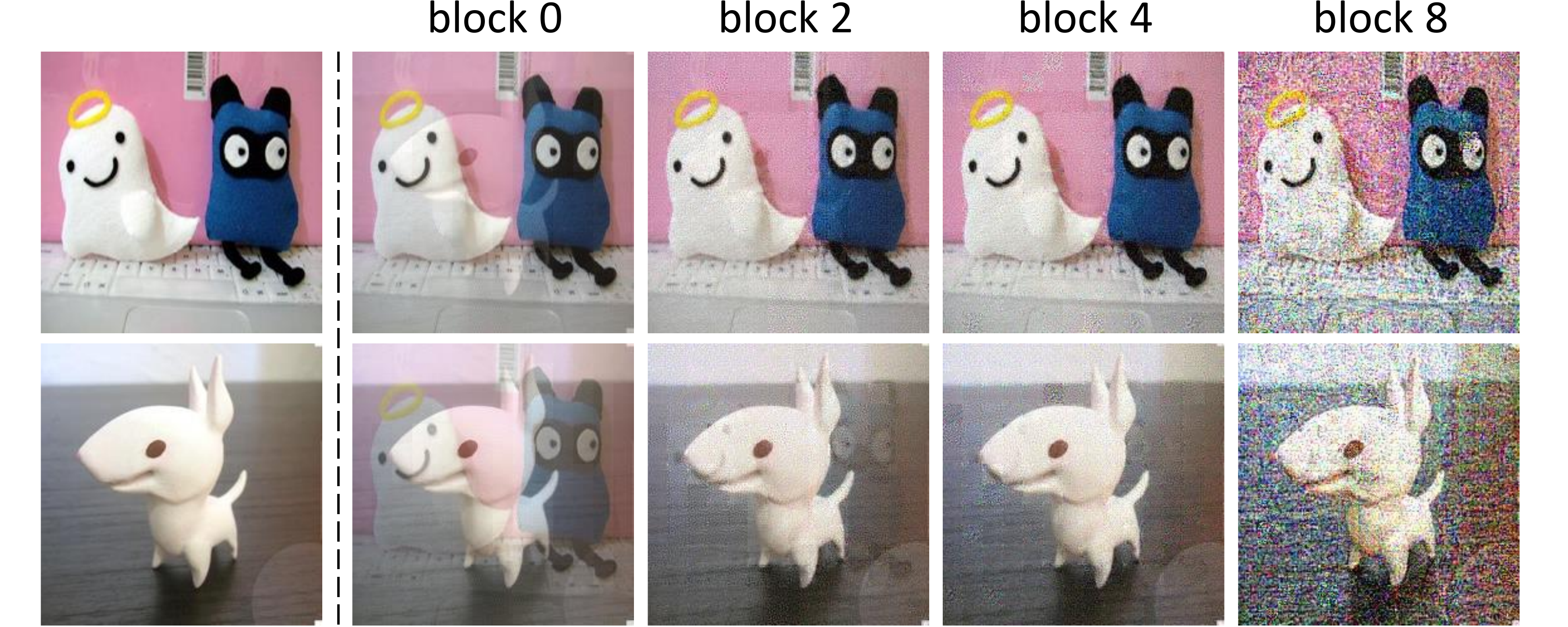}
	\caption{Visualization of perturbation at different layers.}
	\label{fig:visualization}	
	\vspace{-4mm}
\end{figure}

To visualize the perturbed version of a target domain image $\bm{x}$, we initialize a trainable variable $\bm{x}_{vis}$ as $\bm{x}$, and optimize $\bm{x}_{vis}$ to minimize $\Vert \tilde{b}^{l}_x-b^{l}_{xvis}\Vert^2$, where $\tilde{b}^{l}_x$ is the perturbed token sequence of $\bm{x}$ and $b^{l}_{xvis}$ is the corresponding token sequence of $\bm{x}_{vis}$. Then $\bm{x}_{vis}$ gives us an idea on how the perturbation in the latent space reflects on the raw input images. Figure~\ref{fig:visualization} visualizes perturbed version of two images when adding perturbation to different transformer blocks. For shallow layers, an effect of blending with the other image can be observed. However, for deep layers, this effect is less noticeable due to highly non-linear transformation of the network. This also indicates the complementary in using multi-layer perturbation.

\begin{figure}[t]
	\centering
	\begin{subfigure}{0.49\linewidth}
		\includegraphics[width=\linewidth]{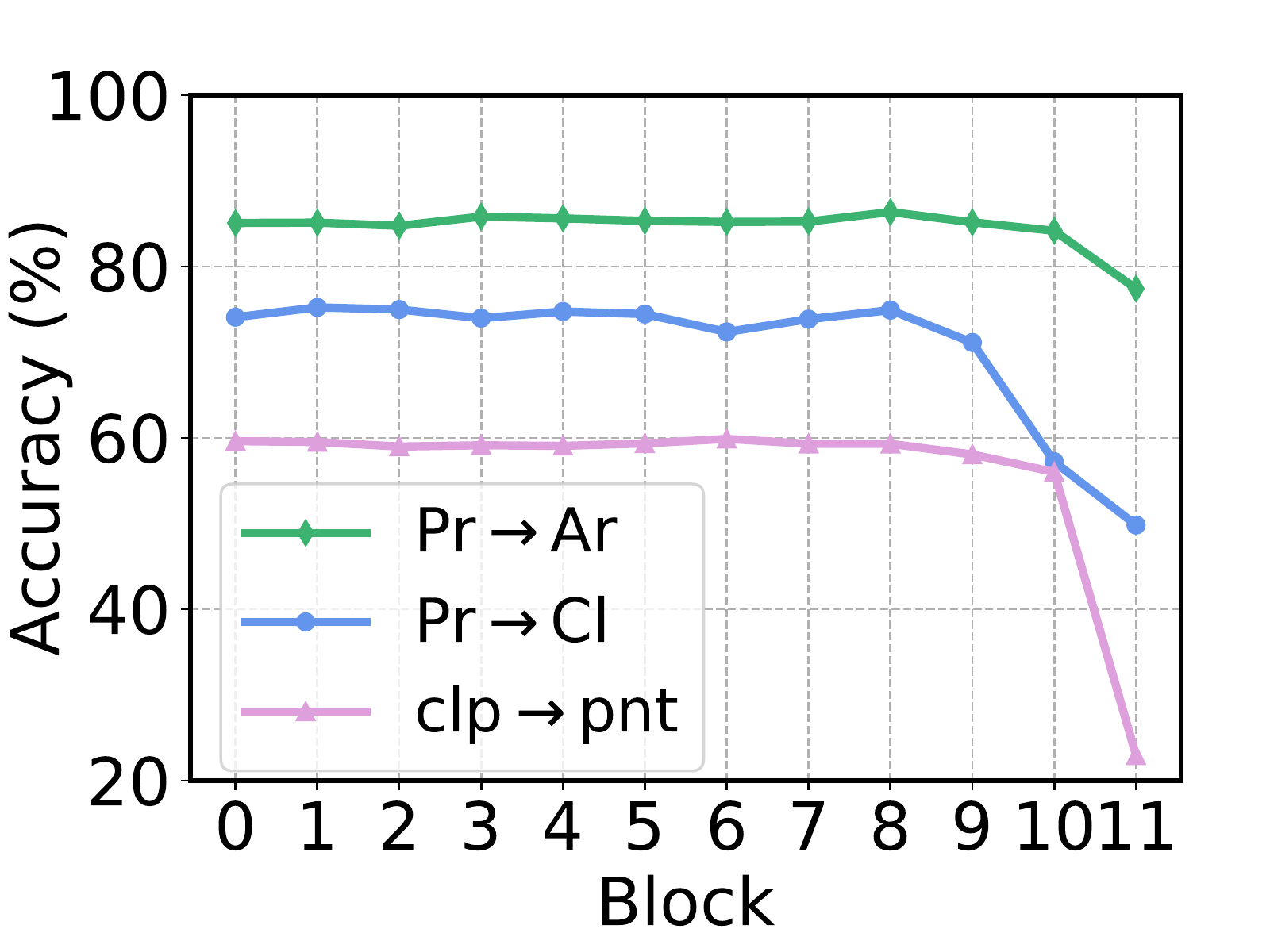}	
		\caption{Perturbation at different layer$^\dagger$}
		\label{fig:abl:layer}
	\end{subfigure}
	\begin{subfigure}{0.49\linewidth}
		%\centering
		\includegraphics[width=\linewidth]{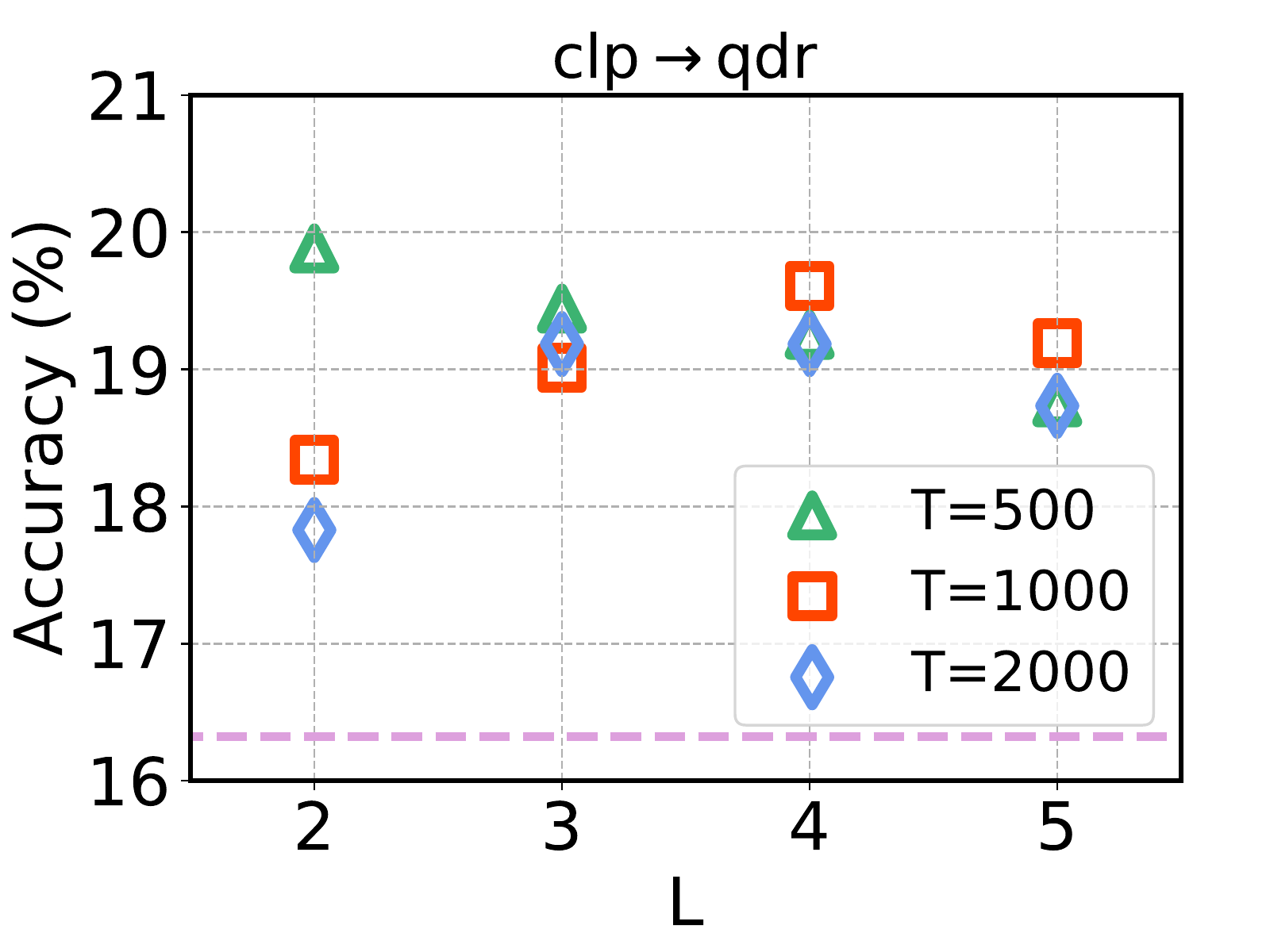}	
		\caption{Safe Training parameters}
		\label{fig:abl:TL}
	\end{subfigure}
	
	\begin{subfigure}{0.49\linewidth}
		\includegraphics[width=\linewidth]{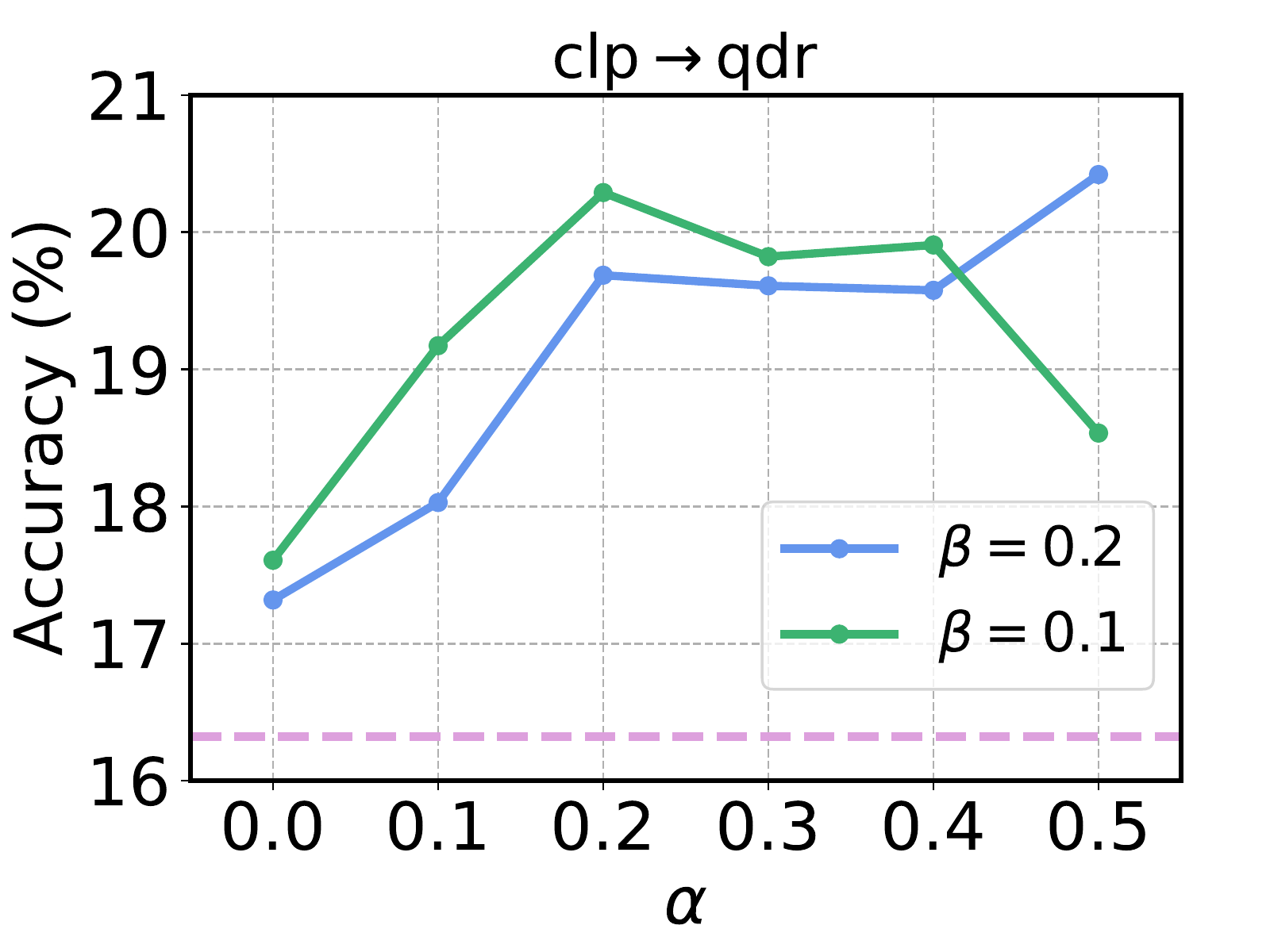}
		\caption{Perturbation scalar}
		\label{fig:abl:alpha}
	\end{subfigure}
	\begin{subfigure}{0.49\linewidth}
		%\centering
		\includegraphics[width=\linewidth]{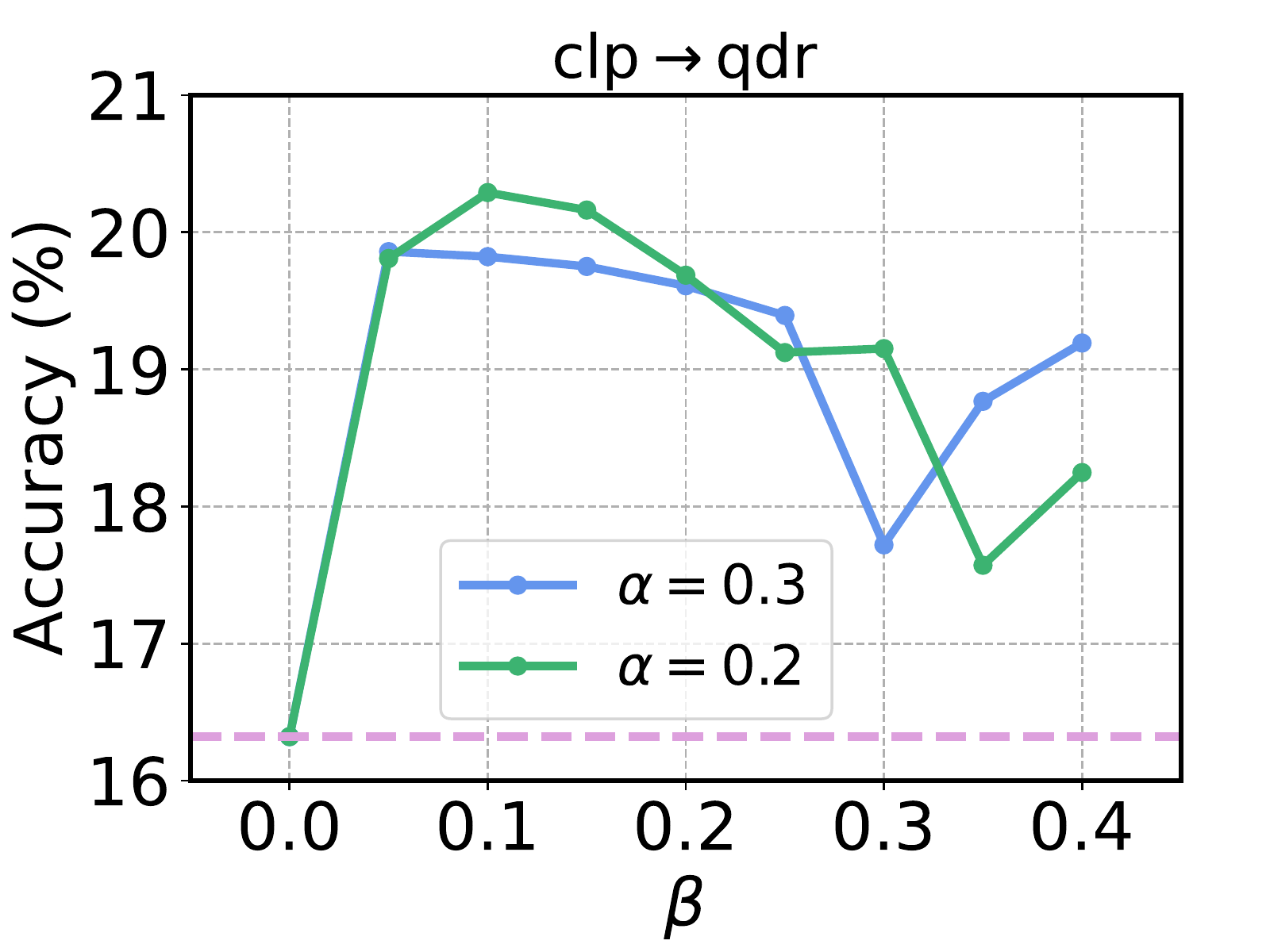}		
		\caption{Self-Refinement loss weight}
		\label{fig:abl:beta}
	\end{subfigure}
	\caption{Plots of ablation studies. Horizontal dash lines indicate baseline accuracies. ($^\dagger$Safe Training not applied)}
	\label{fig:abl}	
\end{figure}

\subsection{Ablation Studies}
\label{sec:abl}
Figure~\ref{fig:abl} presents ablation studies on hyper-parameters. Figure~\ref{fig:abl:layer} plots results of perturbing different layers. Figure~\ref{fig:abl:TL} plots Safe Training with different parameters. $T$ and $L$ affects its granularity. A smaller $T$ implies more quick response. A larger $L$ increases sensitivity but at the risk of more false-positive detections. Many combinations of $T$ and $L$ work well in our method. Figure~\ref{fig:abl:alpha} and~\ref{fig:abl:beta} plots accuracy curves \vs the perturbation scalar $\alpha$ and the self-refinement loss weight $\beta$. Even for obviously unreasonable values like $\alpha=0.5$, Safe Training can still adjust them adaptively to avoid model collapse. When $\alpha=0$, our method still has some gain over baseline. This is due to random dropout operations in the classifier head. 

\section{Conclusion}
In this paper, we propose a novel UDA method named SSRT. It leverages a vision transformer backbone, and uses perturbed target domain data to refine the model. A safe training strategy is developed to avoid model collapse. Experiments on benchmarks show its best performance. 

\textbf{Limitation.} Although we advance the average accuracy on DomainNet to 45.2\%, it is far from saturated. One way is to combine multiple source domains. Another way is to incorporate some meta knowledge about target domains. We plan to extend our study in these directions in the future.

	\appendix
	\renewcommand{\thefigure}{A.\arabic{figure}}
    \setcounter{figure}{0}
    \renewcommand{\thetable}{A.\arabic{table}}
    \setcounter{table}{0}
	
	\section{More Model and Training Details}
	
	Our implementation is based on the \emph{timm} library\footnote{https://github.com/rwightman/pytorch-image-models/blob/master/timm/models/vision\_transformer.py}. We use ViT-B/16~\cite{dosovitskiy2020image} (\emph{vit\_base\_patch16\_224} in \emph{timm}) and ViT-S/16~\cite{dosovitskiy2020image} (\emph{vit\_small\_patch16\_224} in \emph{timm}) as the vision transformer backbones in the paper. Transformer weights are restored from the checkpoints released by official Google JAX implementation\footnote{https://github.com/google-research/vision\_transformer}, which are obtained by first training on ImageNet-21k~\cite{ILSVRC15} and then fine-tuning on Image-1k~\cite{steiner2021augreg,ILSVRC15}. The classifier head consists of a bottleneck module (\texttt{Linear} $\rightarrow$ \texttt{BatchNorm1d} $\rightarrow$ \texttt{ReLU} $\rightarrow$ \texttt{Dropout(0.5)}) and a class predictor (\texttt{Linear} $\rightarrow$ \texttt{ReLU} $\rightarrow$ \texttt{Dropout(0.5)} $\rightarrow$ \texttt{Linear}). The domain discriminator has the same network structure as the class predictor except having only one output. 
	
	During the training procedure, images are first resized to $256\times 256$ pixels, randomly flipped horizontally, and then randomly cropped and resized to $224\times 224$ pixels. The only exception is for VisDA-2017~\cite{peng2017visda}, where center-cropping of size $224\times 224$ is used. During the test procedure, images are first resized to $256\times 256$ pixels and then center-cropped to $224\times 224$ pixels. To train the model, we adopt mini-batch Stochastic Gradient Descent (SGD) with momentum of 0.9. Learning rate is scheduled as $lr = lr_0 * (1 + 1e^{-3}\cdot i) ^ {-0.75}$, where $lr_0$ is initial learning rate, $i$ is training step. The learning rate of parameters of vision transformer backbone is set to be 1/10 of $lr$.  % Complete hyper-parameters used for our experiments are listed in Tab.~\ref{tab:hyper-parameters}. Note that the same hyper-parameters are used for source-only training and baseline methods whenever applicable.  

%	\begin{table}[h]	
%		\caption{Complete list of SSRT hyper-parameters used in the experiments.} 
%		\centering
%		\footnotesize
%		\begin{tabular}{@{\hspace{0.5mm}}p{1.6cm}|p{1.2cm}<{\centering}|p{1.0cm}<{\centering}|p{1.0cm}<{\centering}|p{1.2cm}<{\centering}}
%			\toprule	
%			& Office-31 & Office-Home & VisDA-2017 & DomainNet \\
%			\midrule
%			$\alpha$ & \multicolumn{4}{c}{0.3} \\
%			$\beta$ & \multicolumn{4}{c}{0.2} \\
%			$T$ & \multicolumn{4}{c}{1000} \\
%			$L$ & \multicolumn{4}{c}{4} \\
%			\emph{batch\_size} & \multicolumn{4}{c}{64 (32 source images + 32 target images)} \\
%			\midrule
%			\emph{center\_crop} & False & False & True & False\\
%			% $perturb layers$ & \multicolumn{4}{c}{\{0,4,8\}} \\	 
%			$lr_0$ & 0.001 & 0.004 & 0.002 & 0.004 \\
%%			\emph{bottleneck\_dim} & 1024 & 2048 & 1024 & 1024\\
%			\bottomrule
%		\end{tabular}	\label{tab:hyper-parameters}
%	\end{table}

	\section{More Analysis on Bi-directional Self-Refinement}
	Table~\ref{tab:detach_supp} provides additional results when blocking gradient back-propagation for different variables. Similar to the results listed in the paper (see Tab.~7), allowing gradient back-propagation of the teacher probabilities in KL divergence and $b_x^l$ works better than other variants.

\section{More Analysis on Safe Training}
In our method, we adopt a \textit{Confidence Filter} to remove noisy supervisions. If it not used (\ie, $\epsilon=0$), the performance may deteriorate. Table~\ref{tab:sf_cf} shows that using Safe Training can avoid significant performance drops, making the method much safer.

	\begin{table}[!t]
		\caption{Blocking gradient back-propagation for different variables. Note that $\bm{p}_x$ and $\bm{\tilde{p}}_x$ in the table only refer to the teacher probability in KL divergence. (Safe Training not applied)} 
		\centering
		\footnotesize
		\begin{tabular}{p{1.6cm}|p{0.4cm}<{\centering}p{0.4cm}<{\centering}p{0.4cm}<{\centering}|p{0.8cm}<{\centering}p{0.8cm}<{\centering}p{0.8cm}<{\centering}}	
			\toprule
			& $b^l_x$ & $\bm{p}_x$ & $\bm{\tilde{p}}_x$  &  Cl$\shortrightarrow$Ar &Cl$\shortrightarrow$Pr &  Cl$\shortrightarrow$Rw \\
			\midrule
			$\omega=0$ &  &   & $\times$ &	1.61 & 12.71 &	6.08 \\
			$\omega=1$ &  & $\times$ &  & 81.17 & 85.00 &	87.28  \\
			$\omega\sim\mathcal{B}(0.5)$ &  & $\times$ & $\times$ & 83.68 & 85.69 &	88.04 \\
			\midrule
			$\omega\sim\mathcal{B}(0.5)$ & $\times$ &  &  &	84.55 & 87.27 & 89.49 \\
			$\omega\sim\mathcal{B}(0.5)$ &  &  &  & 85.21 & 87.88 &	89.58 \\
			\bottomrule
		\end{tabular}
		\label{tab:detach_supp}
	\end{table}
	
	\begin{figure}[!t]
		\centering
		\begin{subfigure}{0.49\linewidth}
			\includegraphics[width=\linewidth]{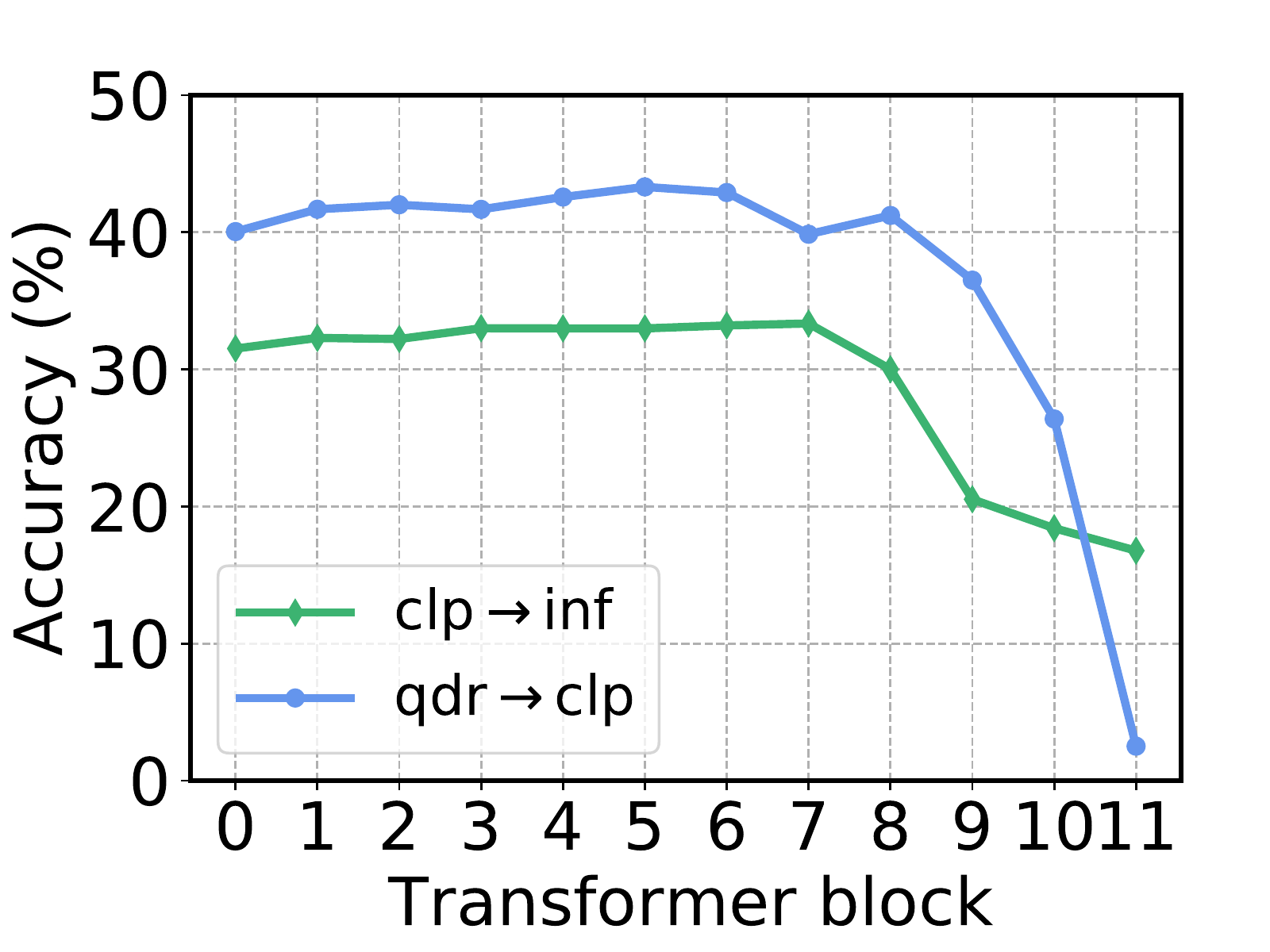}	
			%\caption{Perturbation at different layer$^\dagger$}
			\label{fig:abl_supp:layer}
		\end{subfigure}
		\begin{subfigure}{0.49\linewidth}
			%\centering
			\includegraphics[width=\linewidth]{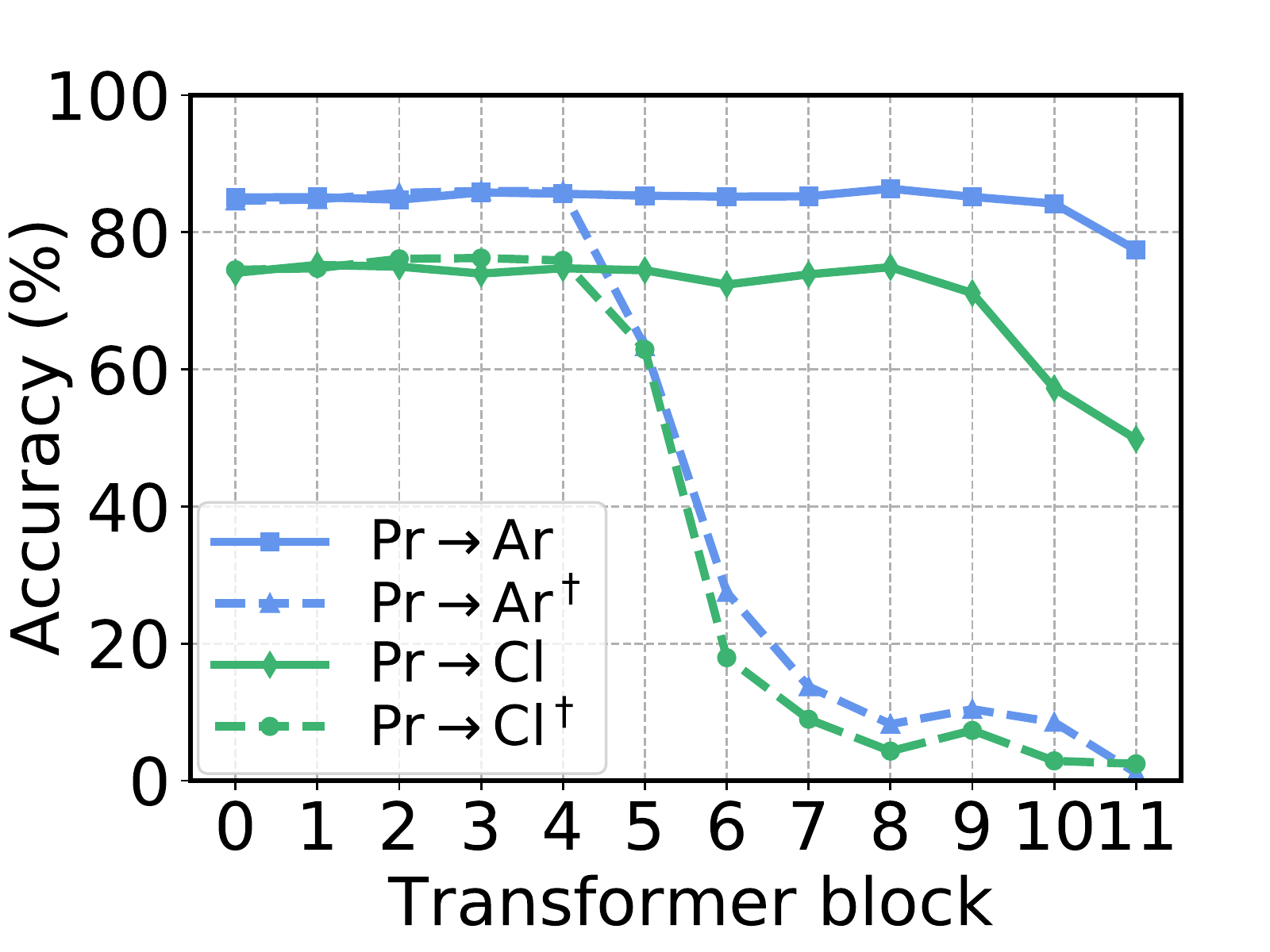}	
			%\caption{Safe Training parameters}
			\label{fig:abl_supp:detach}
		\end{subfigure}	
		\vspace{-4mm}
		\caption{Perturbation at different layer. $^\dagger$No gradient back-propagation for $b_x^l$.}
		\vspace{-2mm}
		\label{fig:abl_supp}
	\end{figure}

	\section{More Analysis on Multi-layer Perturbation}
	Figure~\ref{fig:abl_supp} provides additional results when adding the same amount of perturbation to each layer while not using safe training. As can be seen in the left figure, the best layer to apply perturbation varies across tasks. Besides, a layer that works for one task may fail on others. To see the importance of allowing gradient back-propagation for $b^l_x$ (see Sec.~3.3 and Sec.~3.4 in the paper), the right figure shows that the model collapses when add perturbation to relatively deep layers while blocking the gradients of $b^l_x$.
	
	Table~\ref{tab:domainnet_supp_small} includes comparison results when adding the perturbation to raw input or a single layer (\{0\} or \{4\} or \{8\}) in our proposed SSRT method. As can be seen, perturbing raw input performs similarly to perturbing the \mbox{0-th} transformer block. Besides, perturbing any single layer degrades the performance on some adaptations tasks. In contrast, multi-layer perturbation combines their merits and obtains the best results.
	
		\begin{figure*}[!t]
		\begin{minipage}[t]{0.31\linewidth}
			\centering
			\centering
			\begin{subfigure}{\linewidth}
				\includegraphics[width=\linewidth]{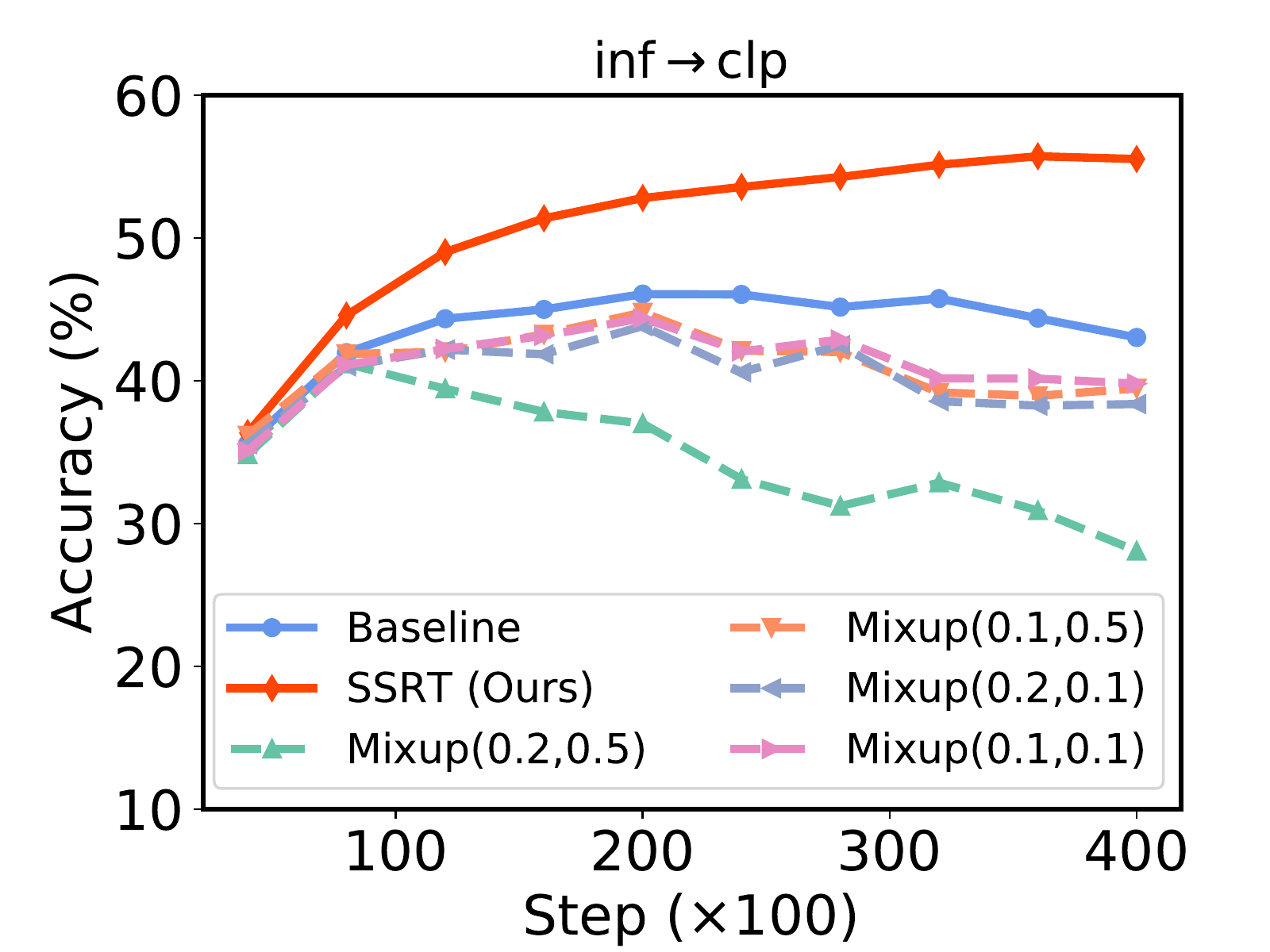}	
			\end{subfigure}
			\begin{subfigure}{\linewidth}
				\includegraphics[width=\linewidth]{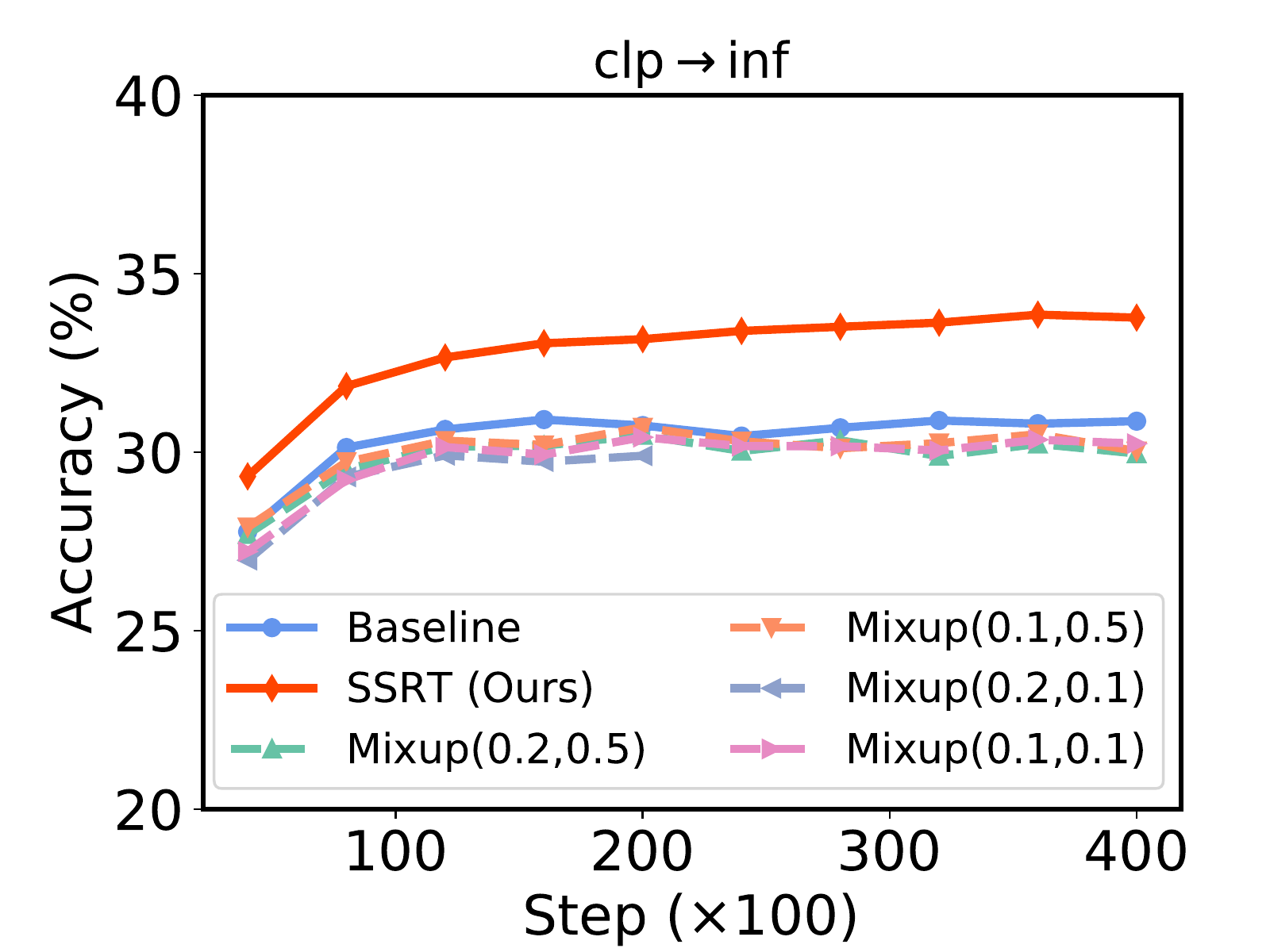}	
			\end{subfigure}
			\caption{Mixup with different hyper-parameters. The legend for Mixup is formed as Mixup($\beta$,$\alpha_{\lambda}$).}
			\label{fig:mixup}	
		\end{minipage}
		\hfill
		\begin{minipage}[t]{0.667\linewidth}
			\centering
			\begin{subfigure}{\linewidth}
				\vspace{1mm}
				\includegraphics[width=\linewidth]{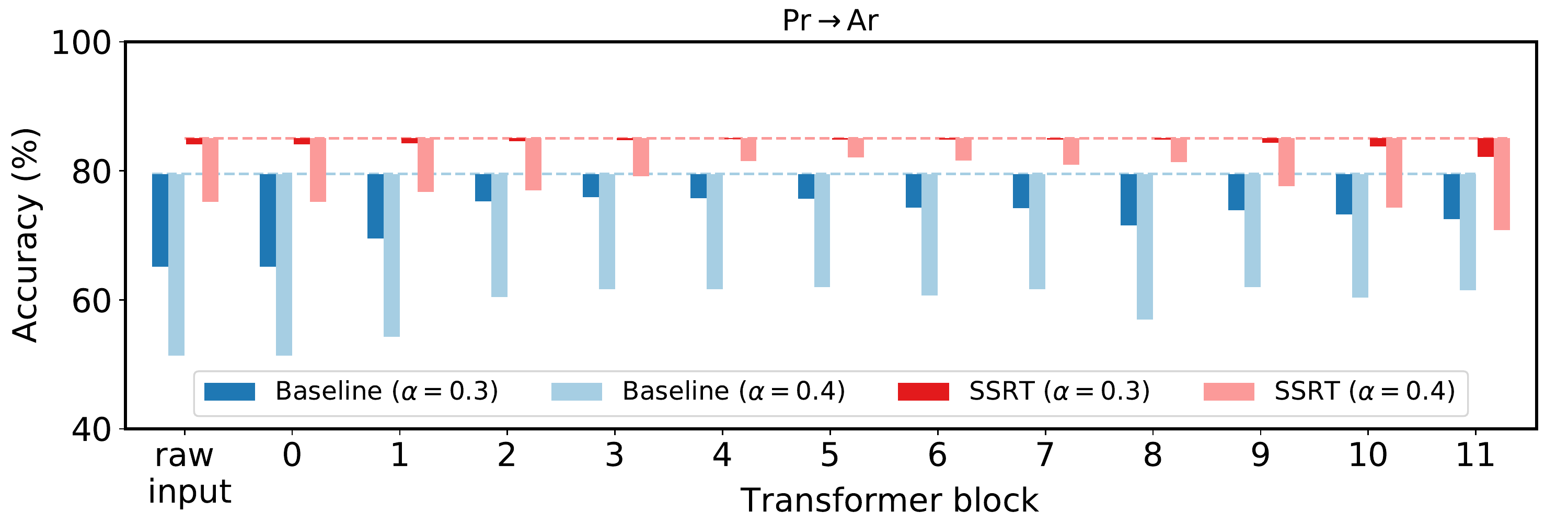}	
				%\label{fig:robust_supp}
				\vspace{-2mm}
			\end{subfigure}		
			\begin{subfigure}{\linewidth}
				\vspace{1mm}
				\includegraphics[width=\linewidth]{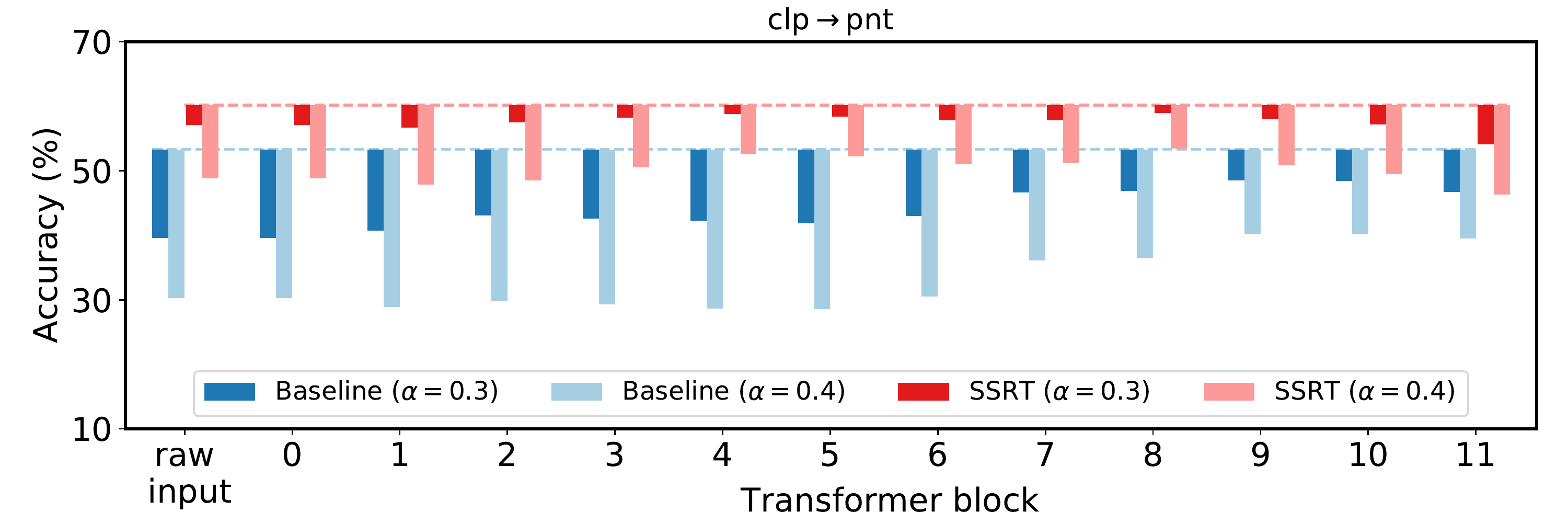}	
			\end{subfigure}
			\caption{Analysis of model's robustness. The dashlines indicate true test accuracy on the target domain data. The bars show decreases of accuracies when adding perturbations to different layers during the test procedure.}
			\label{fig:robust_supp}	
		\end{minipage}
		\vspace{2mm}
	\end{figure*}

	\begin{table*}[!t]
		\caption{Accuracies (\%) on \textbf{DomainNet}. In each sub-table, the column-wise means source domain and the row-wise means target domain. ``-S/B" indicates ViT-small/base backbones, respectively. }
		\scriptsize
		\centering	
		\scalebox{0.98}{
		\input{Table_DomainNet_supp_small.latex}		}
		\label{tab:domainnet_supp_small}
	\end{table*}

	\begin{table*}[t]
			\caption{Accuracies (\%) on \textbf{Office-Home}.}
			\footnotesize
			\centering
			\begin{tabular}{p{1.8cm}p{0.74cm}<{\centering}p{0.74cm}<{\centering}p{0.74cm}<{\centering}p{0.74cm}<{\centering}p{0.74cm}<{\centering}p{0.74cm}<{\centering}p{0.74cm}<{\centering}p{0.74cm}<{\centering}p{0.74cm}<{\centering}p{0.74cm}<{\centering}p{0.74cm}<{\centering}p{0.74cm}<{\centering}>{\columncolor{tbgray}}p{0.74cm}<{\centering}}
				\toprule
				Method & Ar$\shortrightarrow$Cl & Ar$\shortrightarrow$Pr & Ar$\shortrightarrow$Rw &  Cl$\shortrightarrow$Ar & Cl$\shortrightarrow$Pr & Cl$\shortrightarrow$Rw &  Pr$\shortrightarrow$Ar & Pr$\shortrightarrow$Cl & Pr$\shortrightarrow$Rw &      
				Rw$\shortrightarrow$Ar & Rw$\shortrightarrow$Cl & Rw$\shortrightarrow$Pr & Avg.   \\ 	
				\midrule
				Baseline-B & 66.96 &	85.74 &	88.07 &	80.06 &	84.12 &	86.67 &	79.52 &	67.03 &	89.44 &	83.64 &	70.15 &	91.17 &	81.05 \\
				Mixup-B~\cite{zhang2017mixup} & 71.32 &	86.66 &	88.82 &	82.45 &	84.79 &	87.58 &	82.90 &	71.68 &	90.77 &	85.46 &	74.36 &	91.37 &	83.18 \\
				VAT-B~\cite{miyato2018virtual} & 	71.52 &	\HL{89.39} &	90.48 &	\HL{86.11} &	\HL{88.53} &	89.33 &	84.59 &	72.23 &	90.84 &	\HL{86.61} &	72.83 &	\HL{92.48} &	84.58 \\			
				SSRT-B (ours)& \HL{75.17} &	{88.98} &	\HL{91.09} &	{85.13} &	88.29 &	\HL{89.95} &	\HL{85.04} &	\HL{74.23} &	\HL{91.26} &	{85.70} &	\HL{78.58} &	91.78 &	\HL{85.43} \\
				\bottomrule
			\end{tabular} 
			\label{tab:officehome_supp}
 \end{table*}
 
 \begin{table*}[t]
			\caption{Accuracies (\%) on \textbf{VisDA-2017}.}
			\footnotesize
			\centering
			\begin{tabular}{p{1.9cm}p{0.74cm}<{\centering}p{0.74cm}<{\centering}p{0.74cm}<{\centering}p{0.74cm}<{\centering}p{0.74cm}<{\centering}p{0.74cm}<{\centering}p{0.74cm}<{\centering}p{0.74cm}<{\centering}p{0.74cm}<{\centering}p{0.74cm}<{\centering}p{0.74cm}<{\centering}p{0.74cm}<{\centering}>{\columncolor{tbgray}}p{0.74cm}<{\centering}}
				\toprule
				Method  &  plane & bcycl & bus & car & horse & knife & mcycl & person & plant & sktbrd & train & truck & Avg.  \\ \midrule			
				Baseline-B & 98.55 &	82.59 &	85.97 &	57.07 &	94.93 &	97.20 &	94.58 &	76.68 &	92.11 &	96.54 &	94.31 &	52.24 &	85.23 \\
				Mixup-B~\cite{zhang2017mixup} &	98.88 &	86.56 &	88.64 &	72.32 &	98.06 &	98.07 &	95.91 &	\HL{83.00} &	94.09 &	98.07 &	94.55 &	50.36 &	88.21 \\
				VAT-B~\cite{miyato2018virtual} & \HL{99.15} &	\HL{87.71} &	\HL{90.85} &	67.81 &	\HL{98.81} &	98.17 &	\HL{97.57} &	76.65 &	92.88 &	\HL{98.73} &	\HL{96.27} &	\HL{57.37} &	88.50 \\
				SSRT-B (ours)& {98.93} &	87.60 &	{89.10} &	\HL{84.77} &	98.34 &	\HL{98.70} &	96.27 &	{81.08} &	\HL{94.86} &	{97.90} &	94.50 &	43.13 &	\HL{88.76} \\	
				\bottomrule
			\end{tabular}  	
			\label{tab:visda_supp}
	\end{table*}

	\section{Analysis on Model's Robustness}	
	In our proposed SSRT, we use perturbed target domain data to refine the model during the training procedure. In this section, we provide analysis on model's robustness against perturbation during the test procedure. For each testing target domain data, we follow the same way as described in the paper to add a random offset to its latent token sequence, and use the perturbed token sequence to make prediction. To analyze model's robustness against perturbation at different layers, we add perturbation to different transformer block as well as the raw input. The perturbation magnitude is controlled by a scalar $\alpha$ as used in the paper. Figure~\ref{fig:robust_supp} shows results (averaged over 6 random runs) on $Pr\rightarrow Ar$ and $clp\rightarrow pnt$. As can be seen, our method is more robust than Baseline. Even when adding a larger amount of perturbation ($\alpha=0.4$) than seen during training, SSRT incurs less accuracy decrease.

		\begin{table}[t]
		\caption{Comparisons with SSL methods. $X^\dagger$ means averaged over all 5 tasks with $X$ being the target domain.} 
		\centering
		\footnotesize
		\renewcommand\arraystretch{1.2}
		\scalebox{0.88}{
		\begin{tabular}{@{}p{1.3cm}>{\columncolor{tbgray}}p{0.56cm}<{\centering}>{\columncolor{tbgray}}p{0.78cm}<{\centering}>{\columncolor{tbgray}}p{0.78cm}<{\centering}p{0.32cm}<{\centering}p{0.32cm}<{\centering}p{0.32cm}<{\centering}p{0.32cm}<{\centering}p{0.32cm}<{\centering}p{0.32cm}<{\centering}}	
			\toprule
			& Office-Home & VisDA & Domain-Net &  clp$^\dagger$ &	inf$^\dagger$ & pnt$^\dagger$  &	qdr$^\dagger$  &	rel$^\dagger$	 & skt$^\dagger$ \\
			\midrule
			Baseline-B & 81.1 & 85.2 & 38.5 &
			50.6 &	25.6 &	44.9 &	11.6 &	57.0 &	41.5
			\\ 	
			Mixup-B & 83.2 & 	88.2 & -- & --& --& --& --& --& -- \\
			VAT-B &  84.1 & 88.5 & 41.1 & 54.8 &	27.6 &	48.3 &	12.5 &	58.4 &	45.0\\		
			SSRT-B & 85.4 & 88.8 & 45.2 & 60.0 &	28.2 &	53.3 &	13.7 &	65.3 &	50.4 \\		
			\bottomrule
		\end{tabular} }
		\label{tab:ssl}
	\end{table}
	
	\begin{table}[t]
	\caption{Accuracies (\%) without Confidence Filter. ($^\dagger$Safe Training not applied)} 
	\centering
	\footnotesize
	\begin{tabular}{@{\hspace{0.5mm}}p{1.3cm}p{0.7cm}<{\centering}p{0.7cm}<{\centering}p{0.7cm}<{\centering}p{0.7cm}<{\centering}p{0.7cm}<{\centering}p{0.7cm}<{\centering}}	
		\toprule
		&  Cl$\shortrightarrow$Ar &Cl$\shortrightarrow$Pr &  Cl$\shortrightarrow$Rw & Pr$\shortrightarrow$Ar & Pr$\shortrightarrow$Cl  & Pr$\shortrightarrow$Rw  \\
		\midrule
		Baseline-B & 80.06 & 84.12 & 86.67 & 79.52 & 67.03 & 89.44 \\
		SSRT-B$^\dagger$ & 59.33 &	86.98 &	89.74 &	73.92 &	20.30 &	90.59\\
		SSRT-B & 84.51 &	86.98 &	89.30 &	82.65 &	67.79 &	91.16 \\
		\bottomrule
	\end{tabular}
	\label{tab:sf_cf}
\end{table}
	
	\section{Comparison with SSL methods}
	Since Unsupervised Domain Adaptation (UDA) is closely related to Semi-Supervised Learning (SSL), in this section, we compare our method with two representative techniques in SSL, \ie, \emph{Mixup}~\cite{zhang2017mixup} and \emph{VAT}~\cite{miyato2018virtual}. 
	
	Mixup regularizes the model to predict linearly between samples. Specifically, let $\bm{x}_1$ and $\bm{x}_2$ be two target domain data, $p_1=h(\bm{x}_1)$ and $p_2=h(\bm{x}_2)$ be the corresponding model predictions, Mixup first interpolates between two samples by 
	\begin{equation}
		\lambda \sim Beta(\alpha_{\lambda}, \alpha_{\lambda})
	\end{equation}
	\begin{equation}
		\bm{x}^{\prime} = \lambda \bm{x}_1 + (1-\lambda) \bm{x}_2
	\end{equation}
	\begin{equation}
		p^{\prime} = \lambda p_1 + (1-\lambda) p_2
	\end{equation}
	Its loss function is 
	\begin{equation}
		\mathcal{L}_{\rm mixup}=\mathbb{E}_{\bm{x}_1,\bm{x}_2\sim \mathcal{D}_{t}} \| h(\bm{x}^{\prime})-p^{\prime}\|^2
	\end{equation}
	
	VAT enforces the model to predict consistently within the norm-ball neighborhood of each target data $\bm{x}$. Its loss function is
	\begin{equation}
		\mathcal{L}_{\rm VAT}=\mathbb{E}_{\bm{x}\sim \mathcal{D}_{t}}  \bigg[ \max_{\|\bm{r}\|\leq \rho}\mathrm{D}_{KL}\left(h(\bm{x})\| h(\bm{x}+\bm{r})\right) \bigg]
	\end{equation}
	
	We use $\mathcal{L}_{\rm mixup}$ and $\mathcal{L}_{\rm VAT}$ as the $\mathcal{L}_{\rm tgt}$ in our objective function. The trade-off parameter $\beta$ is set to be 0.2 for both, same as used in our method. For Mixup, $\alpha_{\lambda}$ is set to be 0.5. We linearly ramp up $\beta$ to its maximum value over 1/4 of all training steps as used in~\cite{tarvainen2017mean,berthelot2019mixmatch}. Instead of interpolating probabilities, we interpolate unnormalized logits, as it is shown to perform slightly better. For VAT, $\rho$ is set to be 100. Both two techniques are applied to the raw input images. 
	
	Table~\ref{tab:ssl} presents results on three benchmarks using ViT-base backbone. Detailed numbers can be found in Tables~\ref{tab:domainnet_supp_small}-\ref{tab:visda_supp}. On Office-Home~\cite{venkateswara2017deep} and VisDA-2017~\cite{peng2017visda}, Mixup and VAT perform better than Baseline-B, and slightly worse than ours. On DomainNet~\cite{peng2019moment}, VAT still works. However, for Mixup, although we tried different hyper-parameters, it is still inferior to Baseline-B. Figure~\ref{fig:mixup} shows two adaptations tasks where Mixup fails.

%	\begin{table*}[h]
%		\caption{Accuracies (\%) on \textbf{DomainNet}. In each sub-table, the column-wise means source domain and the row-wise means target domain.}
%		\scriptsize
%		\centering	
%		\input{Table_DomainNet_small.latex}	
%		\label{tab:domainnet_small}
%	\end{table*}

\section{Results with ViT-small Backbone}
	\label{sec:vitsmall}
ViT-small is a smaller version of ViT-base by halving the number of Self-Attention Heads and token embedding dimension of ViT-base. It has fewer parameters ($\sim$22M params) than ResNet-101 ($\sim$45M params). We empirically found that it convergences much slower than ViT-base, so we double the maximum training iterations. An alternative is to pretrain the model on the source data first and then adapt it to the target data. As can be seen from Tab.~\ref{tab:domainnet_supp_small}, our proposed SSRT-S achieves +5.1\% higher accuracy than MDD+SCDA (ResNet-101 backbone) on DomainNet, despite that ViT-small has fewer parameters than ResNet-101.

%%%%%%%%% REFERENCES
{\small
\bibliographystyle{ieee_fullname}
\bibliography{paper}
}

\end{document}

%% file: Table_DomainNet_deit.latex
\setlength{\tabcolsep}{4pt}
\begin{tabular}{|p{1.2cm}<{\centering}|@{\hspace{0.05cm}}p{0.55cm}<{\centering}@{\hspace{0.05cm}}p{0.55cm}<{\centering}@{\hspace{0.05cm}}p{0.55cm}<{\centering}@{\hspace{0.05cm}}p{0.55cm}<{\centering}@{\hspace{0.05cm}}p{0.55cm}<{\centering}@{\hspace{0.05cm}}p{0.55cm}<{\centering}@{\hspace{0.2cm}}@{\hspace{0.0cm}}p{0.38cm}<{\centering}||@{\hspace{0.06cm}}p{1.2cm}<{\centering}|@{\hspace{0.05cm}}p{0.55cm}<{\centering}@{\hspace{0.05cm}}p{0.55cm}<{\centering}@{\hspace{0.05cm}}p{0.55cm}<{\centering}@{\hspace{0.05cm}}p{0.55cm}<{\centering}@{\hspace{0.05cm}}p{0.55cm}<{\centering}@{\hspace{0.05cm}}p{0.55cm}<{\centering}@{\hspace{0.2cm}}@{\hspace{0.0cm}}p{0.38cm}<{\centering}||p{1.2cm}<{\centering}|@{\hspace{0.05cm}}p{0.55cm}<{\centering}@{\hspace{0.05cm}}p{0.55cm}<{\centering}@{\hspace{0.05cm}}p{0.55cm}<{\centering}@{\hspace{0.05cm}}p{0.55cm}<{\centering}@{\hspace{0.05cm}}p{0.55cm}<{\centering}@{\hspace{0.05cm}}p{0.55cm}<{\centering}@{\hspace{0.2cm}}@{\hspace{0.0cm}}p{0.38cm}<{\centering}|}
\hline  \tabincell{c}{\textbf{ResNet-}\\\textbf{101}~\cite{he2016deep}}  & clp & inf & pnt & qdr & rel & skt & Avg. &   \tabincell{c}{\textbf{MIMTFL}\\ \cite{gao2020reducing}}  & clp & inf & pnt & qdr & rel & skt & Avg. &  \textbf{CDAN}~\cite{long2018conditional} & clp & inf & pnt & qdr & rel & skt & Avg.  \\\hline clp & -  &  19.3 &  37.5 &  11.1 &  52.2 &  41.0 & 32.2 & clp & -  &  15.1 &  35.6 &  10.7 &  51.5 &  43.1 & 31.2 & clp & -  &  20.4 &  36.6 &  9.0 &  50.7 &  42.3 & 31.8  \\inf &  30.2 & -  &  31.2 &  3.6 &  44.0 &  27.9 & 27.4 & inf &  32.1 & -  &  31.0 &  2.9 &  48.5 &  31.0 & 29.1 & inf &  27.5 & -  &  25.7 &  1.8 &  34.7 &  20.1 & 22.0  \\pnt &  39.6 &  18.7 & -  &  4.9 &  54.5 &  36.3 & 30.8 & pnt &  40.1 &  14.7 & -  &  4.2 &  55.4 &  36.8 & 30.2 & pnt &  42.6 &  20.0 & -  &  2.5 &  55.6 &  38.5 & 31.8  \\qdr &  7.0 &  0.9 &  1.4 & -  &  4.1 &  8.3 & 4.3 & qdr &  18.8 &  3.1 &  5.0 & -  &  16.0 &  13.8 & 11.3 & qdr &  21.0 &  4.5 &  8.1 & -  &  14.3 &  15.7 & 12.7  \\rel &  48.4 &  22.2 &  49.4 &  6.4 & -  &  38.8 & 33.0 & rel &  48.5 &  19.0 &  47.6 &  5.8 & -  &  39.4 & 32.1 & rel &  51.9 &  23.3 &  50.4 &  5.4 & -  &  41.4 & 34.5  \\skt &  46.9 &  15.4 &  37.0 &  10.9 &  47.0 & -  & 31.4 & skt &  51.7 &  16.5 &  40.3 &  12.3 &  53.5 & -  & 34.9 & skt &  50.8 &  20.3 &  43.0 &  2.9 &  50.8 & -  & 33.6  \\Avg. & 34.4 & 15.3 & 31.3 & 7.4 & 40.4 & 30.5 & \cellcolor{tbgray}26.6 & Avg. & 38.2 & 13.7 & 31.9 & 7.2 & 45.0 & 32.8 & \cellcolor{tbgray}28.1 & Avg. & 38.8 & 17.7 & 32.8 & 4.3 & 41.2 & 31.6 & \cellcolor{tbgray}27.7\\ \hline \hline   \tabincell{c}{\textbf{MDD+}\\\textbf{SCDA}~\cite{li2021semantic}}  & clp & inf & pnt & qdr & rel & skt & Avg. &  \tabincell{c}{\textbf{CD-}\\\textbf{Trans$^*$}\cite{xu2021cdtrans}}  & clp & inf & pnt & qdr & rel & skt & Avg. &  \textbf{ViT-B} \cite{dosovitskiy2020image}  & clp & inf & pnt & qdr & rel & skt & Avg.  \\\hline clp & -  &  20.4 &  43.3 &  15.2 &  59.3 &  46.5 & 36.9 & clp & -  &  27.9 &  57.6 &  27.9 &  73.0 &  58.8 & 49.0 & clp & -  &  27.2 &  53.1 &  13.2 &  71.2 &  53.3 & 43.6  \\inf &  32.7 & -  &  34.5 &  6.3 &  47.6 &  29.2 & 30.1 & inf &  58.6 & -  &  53.4 &  9.6 &  71.1 &  47.6 & 48.1 & inf &  51.4 & -  &  49.3 &  4.0 &  66.3 &  41.1 & 42.4  \\pnt &  46.4 &  19.9 & -  &  8.1 &  58.8 &  42.9 & 35.2 & pnt &  60.7 &  24.0 & -  &  13.0 &  69.8 &  49.6 & 43.4 & pnt &  53.1 &  25.6 & -  &  4.8 &  70.0 &  41.8 & 39.1  \\qdr &  31.1 &  6.6 &  18.0 & -  &  28.8 &  22.0 & 21.3 & qdr &  2.9 &  0.4 &  0.3 & -  &  0.7 &  4.7 & 1.8 & qdr &  30.5 &  4.5 &  16.0 & -  &  27.0 &  19.3 & 19.5  \\rel &  55.5 &  23.7 &  52.9 &  9.5 & -  &  45.2 & 37.4 & rel &  49.3 &  18.7 &  47.8 &  9.4 & -  &  33.5 & 31.7 & rel &  58.4 &  29.0 &  60.0 &  6.0 & -  &  45.8 & 39.9  \\skt &  55.8 &  20.1 &  46.5 &  15.0 &  56.7 & -  & 38.8 & skt &  66.8 &  23.7 &  54.6 &  27.5 &  68.0 & -  & 48.1 & skt &  63.9 &  23.8 &  52.3 &  14.4 &  67.4 & -  & 44.4  \\Avg. & 44.3 & 18.1 & 39.0 & 10.8 & 50.2 & 37.2 & \cellcolor{tbgray}33.3 & Avg. & 47.7 & 18.9 & 42.7 & 17.5 & 56.5 & 38.8 & \cellcolor{tbgray}37.0 & Avg. & 51.5 & 22.0 & 46.1 & 8.5 & 60.4 & 40.3 & \cellcolor{tbgray}38.1\\ \hline \hline  \textbf{Baseline-B}  & clp & inf & pnt & qdr & rel & skt & Avg. &  \textbf{\tabincell{c}{Baseline-B\\+MI}}  & clp & inf & pnt & qdr & rel & skt & Avg. &  \textbf{\tabincell{c}{SSRT-B\\(ours)}}  & clp & inf & pnt & qdr & rel & skt & Avg.  \\\hline clp & -  &  30.9 &  53.3 &  16.3 &  72.7 &  55.4 & 45.7 & clp & -  &  30.5 &  55.8 &  18.1 &  74.7 &  57.5 & 47.3 & clp & -  &  33.8 &  60.2 &  19.4 &  75.8 &  59.8 & 49.8  \\inf &  43.0 & -  &  40.8 &  7.8 &  56.4 &  35.9 & 36.8 & inf &  53.2 & -  &  52.8 &  9.2 &  68.3 &  45.3 & 45.8 & inf &  55.5 & -  &  54.0 &  9.0 &  68.2 &  44.7 & 46.3  \\pnt &  55.7 &  28.6 & -  &  7.4 &  70.5 &  48.3 & 42.1 & pnt &  56.8 &  27.6 & -  &  7.3 &  70.8 &  49.3 & 42.4 & pnt &  61.7 &  28.5 & -  &  8.4 &  71.4 &  55.2 & 45.0  \\qdr &  25.5 &  5.2 &  9.7 & -  &  15.5 &  17.1 & 14.6 & qdr &  31.6 &  5.1 &  13.3 & -  &  25.3 &  23.0 & 19.6 & qdr &  42.5 &  8.8 &  24.2 & -  &  37.6 &  33.6 & 29.3  \\rel &  62.3 &  32.5 &  62.5 &  8.2 & -  &  50.7 & 43.2 & rel &  65.7 &  32.4 &  63.9 &  6.9 & -  &  51.7 & 44.1 & rel &  69.9 &  37.1 &  66.0 &  10.1 & -  &  58.9 & 48.4  \\skt &  66.4 &  30.6 &  58.0 &  18.1 &  70.1 & -  & 48.6 & skt &  68.9 &  30.6 &  61.0 &  19.3 &  72.9 & -  & 50.5 & skt &  70.6 &  32.8 &  62.2 &  21.7 &  73.2 & -  & 52.1  \\Avg. & 50.6 & 25.6 & 44.9 & 11.6 & 57.0 & 41.5 & \cellcolor{tbgray}38.5 & Avg. & 55.2 & 25.2 & 49.4 & 12.2 & 62.4 & 45.3 & \cellcolor{tbgray}41.6 & Avg. & 60.0 & 28.2 & 53.3 & 13.7 & 65.3 & 50.4 & \cellcolor{tbgray}\HL{45.2}\\ \hline 
\end{tabular}

%% file: Table_DomainNet_supp_small.latex
\setlength{\tabcolsep}{4pt}
\begin{tabular}{|p{1.2cm}<{\centering}|@{\hspace{0.05cm}}p{0.55cm}<{\centering}@{\hspace{0.05cm}}p{0.55cm}<{\centering}@{\hspace{0.05cm}}p{0.55cm}<{\centering}@{\hspace{0.05cm}}p{0.55cm}<{\centering}@{\hspace{0.05cm}}p{0.55cm}<{\centering}@{\hspace{0.05cm}}p{0.55cm}<{\centering}@{\hspace{0.2cm}}@{\hspace{0.0cm}}p{0.38cm}<{\centering}||p{1.2cm}<{\centering}|@{\hspace{0.05cm}}p{0.55cm}<{\centering}@{\hspace{0.05cm}}p{0.55cm}<{\centering}@{\hspace{0.05cm}}p{0.55cm}<{\centering}@{\hspace{0.05cm}}p{0.55cm}<{\centering}@{\hspace{0.05cm}}p{0.55cm}<{\centering}@{\hspace{0.05cm}}p{0.55cm}<{\centering}@{\hspace{0.2cm}}@{\hspace{0.0cm}}p{0.38cm}<{\centering}||p{1.2cm}<{\centering}|@{\hspace{0.05cm}}p{0.55cm}<{\centering}@{\hspace{0.05cm}}p{0.55cm}<{\centering}@{\hspace{0.05cm}}p{0.55cm}<{\centering}@{\hspace{0.05cm}}p{0.55cm}<{\centering}@{\hspace{0.05cm}}p{0.55cm}<{\centering}@{\hspace{0.05cm}}p{0.55cm}<{\centering}@{\hspace{0.2cm}}@{\hspace{0.0cm}}p{0.38cm}<{\centering}|}
\hline  \tabincell{c}{\textbf{MDD+}\\\textbf{SCDA}~\cite{li2021semantic}} & clp & inf & pnt & qdr & rel & skt & Avg. &  \textbf{ViT-B}  & clp & inf & pnt & qdr & rel & skt & Avg. &  \textbf{Baseline-B}  & clp & inf & pnt & qdr & rel & skt & Avg.  \\\hline clp & -  &  20.4 &  43.3 &  15.2 &  59.3 &  46.5 & 36.9 & clp & -  &  27.2 &  53.1 &  13.2 &  71.2 &  53.3 & 43.6 & clp & -  &  30.9 &  53.3 &  16.3 &  72.7 &  55.4 & 45.7  \\inf &  32.7 & -  &  34.5 &  6.3 &  47.6 &  29.2 & 30.1 & inf &  51.4 & -  &  49.3 &  4.0 &  66.3 &  41.1 & 42.4 & inf &  43.0 & -  &  40.8 &  7.8 &  56.4 &  35.9 & 36.8  \\pnt &  46.4 &  19.9 & -  &  8.1 &  58.8 &  42.9 & 35.2 & pnt &  53.1 &  25.6 & -  &  4.8 &  70.0 &  41.8 & 39.1 & pnt &  55.7 &  28.6 & -  &  7.4 &  70.5 &  48.3 & 42.1  \\qdr &  31.1 &  6.6 &  18.0 & -  &  28.8 &  22.0 & 21.3 & qdr &  30.5 &  4.5 &  16.0 & -  &  27.0 &  19.3 & 19.5 & qdr &  25.5 &  5.2 &  9.7 & -  &  15.5 &  17.1 & 14.6  \\rel &  55.5 &  23.7 &  52.9 &  9.5 & -  &  45.2 & 37.4 & rel &  58.4 &  29.0 &  60.0 &  6.0 & -  &  45.8 & 39.9 & rel &  62.3 &  32.5 &  62.5 &  8.2 & -  &  50.7 & 43.2  \\skt &  55.8 &  20.1 &  46.5 &  15.0 &  56.7 & -  & 38.8 & skt &  63.9 &  23.8 &  52.3 &  14.4 &  67.4 & -  & 44.4 & skt &  66.4 &  30.6 &  58.0 &  18.1 &  70.1 & -  & 48.6  \\Avg. & 44.3 & 18.1 & 39.0 & 10.8 & 50.2 & 37.2 & \cellcolor{tbgray}33.3 & Avg. & 51.5 & 22.0 & 46.1 & 8.5 & 60.4 & 40.3 & \cellcolor{tbgray}38.1 & Avg. & 50.6 & 25.6 & 44.9 & 11.6 & 57.0 & 41.5 & \cellcolor{tbgray}38.5\\ \hline \hline  \textbf{VAT-B}\cite{miyato2018virtual}  & clp & inf & pnt & qdr & rel & skt & Avg. &  \textbf{\tabincell{c}{SSRT-B\\\ raw input}}  & clp & inf & pnt & qdr & rel & skt & Avg. &  \textbf{\tabincell{c}{SSRT-B\\\{0\}}}  & clp & inf & pnt & qdr & rel & skt & Avg.  \\\hline clp & -  &  33.1 &  57.1 &  19.5 &  75.8 &  59.8 & 49.0 & clp & -  &  32.7 &  60.0 &  19.0 &  75.3 &  59.8 & 49.3 & clp & -  &  33.2 &  59.7 &  19.6 &  75.3 &  58.7 & 49.3  \\inf &  48.3 & -  &  45.2 &  9.8 &  55.0 &  37.4 & 39.2 & inf &  55.0 & -  &  54.0 &  8.9 &  67.8 &  48.1 & 46.8 & inf &  54.8 & -  &  53.5 &  9.3 &  67.7 &  46.1 & 46.3  \\pnt &  60.0 &  30.9 & -  &  7.9 &  71.1 &  52.6 & 44.5 & pnt &  61.6 &  28.6 & -  &  8.2 &  71.3 &  55.4 & 45.0 & pnt &  61.2 &  29.0 & -  &  7.1 &  71.2 &  55.0 & 44.7  \\qdr &  26.7 &  5.4 &  9.2 & -  &  18.1 &  18.3 & 15.5 & qdr &  36.3 &  6.2 &  16.1 & -  &  32.1 &  31.2 & 24.4 & qdr &  40.8 &  7.0 &  13.2 & -  &  35.4 &  31.1 & 25.5  \\rel &  68.7 &  35.3 &  65.0 &  7.8 & -  &  56.8 & 46.7 & rel &  69.8 &  35.6 &  66.1 &  12.4 & -  &  59.2 & 48.6 & rel &  69.6 &  35.7 &  65.7 &  10.7 & -  &  58.7 & 48.1  \\skt &  70.2 &  33.3 &  65.0 &  17.6 &  72.2 & -  & 51.7 & skt &  70.3 &  30.5 &  62.3 &  20.0 &  73.2 & -  & 51.3 & skt &  69.7 &  32.1 &  62.0 &  19.0 &  72.8 & -  & 51.1  \\Avg. & 54.8 & 27.6 & 48.3 & 12.5 & 58.4 & 45.0 & \cellcolor{tbgray}41.1 & Avg. & 58.6 & 26.7 & 51.7 & 13.7 & 63.9 & 50.8 & \cellcolor{tbgray}44.2 & Avg. & 59.2 & 27.4 & 50.8 & 13.1 & 64.5 & 49.9 & \cellcolor{tbgray}44.2\\ \hline \hline  \textbf{\tabincell{c}{SSRT-B\\\{4\}}}  & clp & inf & pnt & qdr & rel & skt & Avg. &  \textbf{\tabincell{c}{SSRT-B\\\{8\}}}  & clp & inf & pnt & qdr & rel & skt & Avg. &  \textbf{\tabincell{c}{SSRT-B\\\{0,4,8\}}}  & clp & inf & pnt & qdr & rel & skt & Avg.  \\\hline clp & -  &  31.8 &  58.9 &  17.8 &  75.7 &  59.4 & 48.7 & clp & -  &  32.4 &  59.0 &  18.6 &  75.6 &  59.9 & 49.1 & clp & -  &  33.8 &  60.2 &  19.4 &  75.8 &  59.8 & 49.8  \\inf &  53.5 & -  &  50.5 &  8.6 &  67.8 &  47.5 & 45.6 & inf &  55.9 & -  &  54.8 &  7.6 &  68.5 &  48.2 & 47.0 & inf &  55.5 & -  &  54.0 &  9.0 &  68.2 &  44.7 & 46.3  \\pnt &  61.3 &  29.2 & -  &  8.1 &  71.3 &  54.3 & 44.8 & pnt &  61.5 &  27.4 & -  &  8.5 &  71.4 &  54.6 & 44.7 & pnt &  61.7 &  28.5 & -  &  8.4 &  71.4 &  55.2 & 45.0  \\qdr &  42.5 &  7.7 &  17.0 & -  &  23.3 &  33.4 & 24.8 & qdr &  33.6 &  5.7 &  11.3 & -  &  31.4 &  31.8 & 22.7 & qdr &  42.5 &  8.8 &  24.2 & -  &  37.6 &  33.6 & 29.3  \\rel &  68.7 &  36.1 &  65.5 &  8.2 & -  &  57.6 & 47.2 & rel &  69.6 &  36.2 &  65.9 &  6.9 & -  &  58.1 & 47.3 & rel &  69.9 &  37.1 &  66.0 &  10.1 & -  &  58.9 & 48.4  \\skt &  70.1 &  31.8 &  62.2 &  17.7 &  73.1 & -  & 51.0 & skt &  69.9 &  30.9 &  62.3 &  19.8 &  73.3 & -  & 51.2 & skt &  70.6 &  32.8 &  62.2 &  21.7 &  73.2 & -  & 52.1  \\Avg. & 59.2 & 27.3 & 50.8 & 12.1 & 62.2 & 50.4 & \cellcolor{tbgray}43.7 & Avg. & 58.1 & 26.5 & 50.6 & 12.3 & 64.0 & 50.5 & \cellcolor{tbgray}43.7 & Avg. & 60.0 & 28.2 & 53.3 & 13.7 & 65.3 & 50.4 & \cellcolor{tbgray}45.2\\ \hline \hline  \textbf{ViT-S}  & clp & inf & pnt & qdr & rel & skt & Avg. &  \textbf{Baseline-S}  & clp & inf & pnt & qdr & rel & skt & Avg. &  \textbf{\tabincell{c}{\vspace{-1.5mm}\\ SSRT-S \\ \vspace{-1.5mm} }}  & clp & inf & pnt & qdr & rel & skt & Avg.  \\\hline clp & -  &  23.0 &  46.2 &  11.9 &  66.3 &  46.2 & 38.7 & clp & -  &  27.0 &  49.0 &  12.8 &  68.2 &  49.1 & 41.2 & clp & -  &  28.5 &  53.1 &  12.1 &  69.9 &  52.1 & 43.1  \\inf &  42.9 & -  &  42.8 &  3.8 &  62.3 &  33.9 & 37.1 & inf &  41.8 & -  &  43.1 &  2.7 &  63.0 &  33.0 & 36.7 & inf &  47.5 & -  &  49.8 &  1.5 &  64.9 &  39.7 & 40.7  \\pnt &  45.2 &  22.2 & -  &  3.5 &  66.5 &  35.7 & 34.6 & pnt &  48.8 &  25.7 & -  &  3.1 &  67.0 &  40.8 & 37.1 & pnt &  53.0 &  26.5 & -  &  4.4 &  67.3 &  46.7 & 39.6  \\qdr &  19.7 &  3.3 &  7.8 & -  &  14.6 &  12.7 & 11.6 & qdr &  21.8 &  5.8 &  9.6 & -  &  15.3 &  15.2 & 13.5 & qdr &  31.3 &  6.9 &  13.0 & -  &  24.4 &  24.0 & 19.9  \\rel &  50.8 &  24.2 &  54.2 &  4.6 & -  &  37.3 & 34.2 & rel &  54.6 &  28.7 &  57.5 &  3.6 & -  &  41.3 & 37.1 & rel &  60.0 &  31.2 &  60.5 &  4.6 & -  &  48.5 & 41.0  \\skt &  57.2 &  19.5 &  47.1 &  13.9 &  62.5 & -  & 40.0 & skt &  60.9 &  26.2 &  53.9 &  10.6 &  67.5 & -  & 43.8 & skt &  63.8 &  28.6 &  57.0 &  13.7 &  68.7 & -  & 46.4  \\Avg. & 43.1 & 18.5 & 39.6 & 7.5 & 54.4 & 33.2 & \cellcolor{tbgray}32.7 & Avg. & 45.6 & 22.7 & 42.6 & 6.5 & 56.2 & 35.9 & \cellcolor{tbgray}34.9 & Avg. & 51.1 & 24.4 & 46.7 & 7.3 & 59.0 & 42.2 & \cellcolor{tbgray}38.4\\ \hline 
\end{tabular}